\PassOptionsToPackage{table}{xcolor}
\documentclass[11pt]{article}

\usepackage[final]{acl}

\usepackage{times}
\usepackage{latexsym}
\usepackage{xurl}

\usepackage[T1]{fontenc}

\usepackage[utf8]{inputenc}

\usepackage{microtype}

\usepackage{inconsolata}

\usepackage{algorithm}
\usepackage{algpseudocode}
\usepackage{capt-of}
\usepackage{mdframed}
\usepackage{fvextra}

\usepackage{amsfonts}
\usepackage{amssymb}
\usepackage{hyperref}
\usepackage{bm}
\usepackage{multirow}
\usepackage{multicol}

\usepackage{graphicx}
\usepackage{booktabs}
\usepackage{framed}
\usepackage{caption}
\usepackage{subcaption}
\usepackage{xspace}
\usepackage{enumitem}
\usepackage{color,colortbl}
\usepackage{amsthm}
\usepackage{amsmath}
\usepackage{tcolorbox}
\tcbuselibrary{raster}
\usepackage{cleveref}
\usepackage[inkscapelatex=false]{svg}
\usepackage{dashrule}
\usepackage{arydshln}
\usepackage{overpic}
\usepackage{marvosym}
\usepackage{adjustbox}
\usepackage{pifont}

\newcommand{\cdashlinelr}[1]{%
  \noalign{\vskip\aboverulesep
           \global\let\@dashdrawstore\adl@draw
           \global\let\adl@draw\adl@drawiv}
  \cdashline{#1}
  \noalign{\global\let\adl@draw\@dashdrawstore
           \vskip\belowrulesep}}

\makeatletter
\def\@fnsymbol#1{}
\makeatother

\newtheorem{insight}{Finding}

\tcbuselibrary{skins, breakable}

\definecolor{cotred}{RGB}{178,34,34}
\definecolor{spsblue}{RGB}{50,100,180}
\definecolor{lightred}{RGB}{252,242,242}
\definecolor{lightblue}{RGB}{220,235,250}

\newtcolorbox[auto counter]{SummaryBox}[1][]{ 
    enhanced,
    breakable,
    colback=lightblue!80,         
    colframe=black,               
    fonttitle=\bfseries\fontsize{10.1pt}{11.2pt}\selectfont,
    coltitle=white,               
    colbacktitle=black,           
    title={Key Finding \thetcbcounter}, 
    width=\linewidth,             
    arc=3.5mm,                    
    attach boxed title to top left={xshift=2.5mm, yshift=-2.5mm},
    boxed title style={rounded corners, size=small, colframe=black, colback=black},
    top=4mm, bottom=2mm, left=2mm, right=2mm,
}

\title{Efficient Reasoning Exploration via State-Conditioned Latent Steering with Progress Guidance}

\author{
 \textbf{Hengyuan Zhang\textsuperscript{1}},
 \textbf{Chenming Shang\textsuperscript{1}},
 \textbf{Zunhai Su\textsuperscript{1}},
 \textbf{Xiao Liang\textsuperscript{2}},
 \textbf{Hui Shen},
 \textbf{Jing Xiong\textsuperscript{1}},
 \textbf{Dawei Li\textsuperscript{3}},\\
 \textbf{Shiping Yang\textsuperscript{4}},
 \textbf{Kailai Yang\textsuperscript{1,5}}, 
 \textbf{Wei Zhang\textsuperscript{6}},
 \textbf{Ruobing Xie\textsuperscript{6}},
 \textbf{Hayden Kwok-Hay So\textsuperscript{1}}, 
 \textbf{Ngai Wong\textsuperscript{1 \dag}}\thanks{\dag\ Corresponding author.}
\\
 \textsuperscript{1}The University of Hong Kong  \ \
 \textsuperscript{2}University of California, Los Angeles  \ \ 
 \textsuperscript{3}Arizona State University  \ \ \\
 \textsuperscript{4}Simon Fraser University  \ \ 
 \textsuperscript{5}The University of Manchester  \ \ 
 \textsuperscript{6}Tencent 
 \\
\texttt{hengyuan.zhang88@gmail.com }  
}

\begin{document}
\maketitle

\begin{abstract}

Best-of-$N$ is a widely used inference strategy for complex reasoning, whose effectiveness depends on whether sampled candidates can cover diverse and high-quality reasoning paths.
However, post-trained reasoning models often suffer from \emph{exploration collapse}, where independent rollouts repeatedly follow similar reasoning paths and limit the gains from increasing the rollout budget.
Existing methods alleviate this issue by promoting broader exploration, but do not explicitly guide exploration toward continuations that make meaningful progress, resulting in limited exploration efficiency.
To address this, we propose \emph{\underline{S}tate-conditioned \underline{P}rogress-guided \underline{S}teering} (SPS), a training-free latent steering framework.
Specifically, SPS constructs a state-conditioned Direction Bank containing multiple progress-guided steering vectors for different prefix-state regions.
During online inference, SPS retrieves a suitable steering vector based on the current prefix state and applies it at high-uncertainty transitions to guide the next reasoning step toward meaningful progress.
Extensive experiments across multiple model scales and benchmarks demonstrate that SPS consistently outperforms strong baselines.
Further analyses validate the effectiveness of its key designs and offer valuable insights for future research.
The code is available at \href{https://github.com/rattlesnakey/SPS}{the provided link}.
\end{abstract}

\section{Introduction}
\label{sec:intro}
\begin{figure}[!t]
    \centering \centerline{\includegraphics[width=\columnwidth]{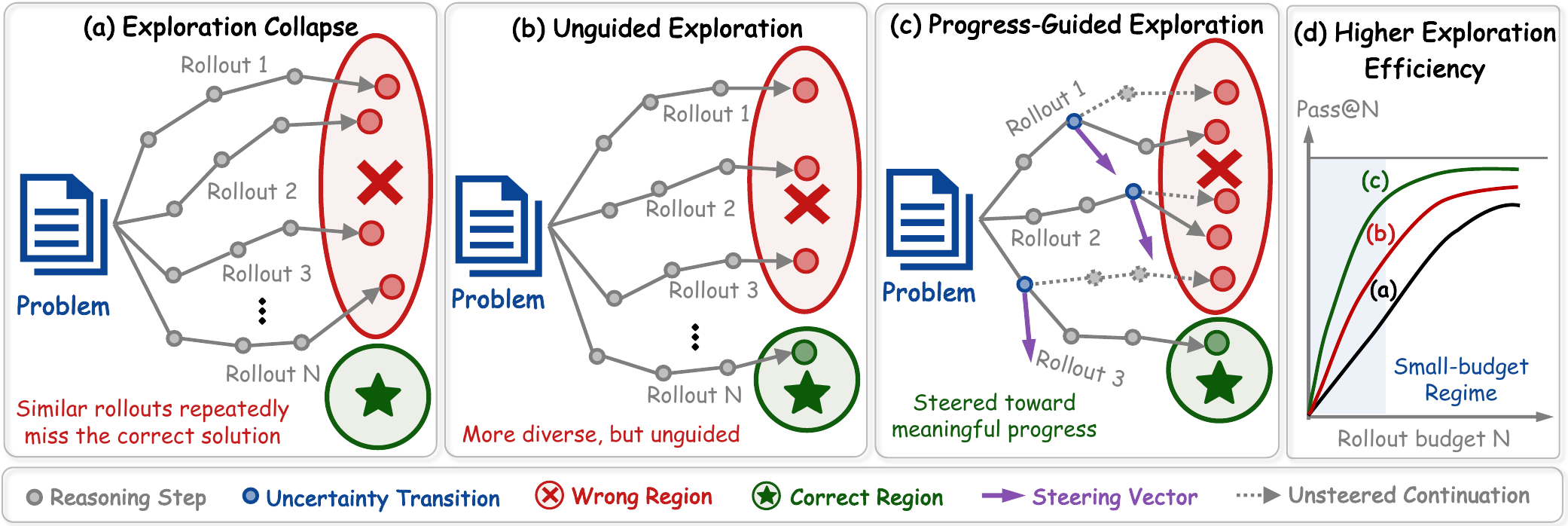}}
    \vspace{-0.1cm}
   \caption{
(a) Exploration collapse causes similar rollouts to repeatedly miss the correct solution.
(b) Unguided exploration increases diversity but may still need many rollouts to find a correct solution.
(c) Progress-guided steering intervenes at high-uncertainty transitions to guide reasoning toward meaningful progress, increasing the chance of finding a correct solution with fewer rollouts.
(d) Higher exploration efficiency means finding a correct solution with fewer rollouts.
}
    \vspace{-0.2cm}
    \label{fig:intro_motivation}
\end{figure}

Recent Large Language Models (LLMs) have demonstrated strong capabilities in complex reasoning~\citep{li2025system,yu2025chain,qwen3,liang2026training}.
A widely used strategy for further improving reasoning performance at inference time is Best-of-$N$, where a model first samples $N$ candidate solutions and a verifier or reward model then selects the best candidate~\citep{lightman2024let,brown2024large}.
Its effectiveness therefore depends heavily on whether the sampled candidates can cover diverse and high-quality reasoning paths. 
However, modern reasoning models typically undergo supervised fine-tuning (SFT) and reinforcement learning (RL) post-training~\citep{qwen3,guo2025deepseek,ace_nemotron}. 
This post-training substantially improves Pass@1 but also induces \emph{exploration collapse}, where independent rollouts repeatedly follow similar reasoning paths, limiting the gains from Best-of-$N$ sampling~\citep{exploration_collapse_1,exploration_collapse_2,exploration_collapse_3}.

Existing work has explored several ways to alleviate exploration collapse. 
One line of work modifies the training procedure to preserve reasoning diversity or encourage exploration~\citep{sft_diverse_1,sft_diverse_2,rl_diverse_1,yang2026modularized,rl_diverse_2,rl_diverse_4}. 
Although effective, these approaches require additional training and often involve specialized data, loss functions, or reward designs, increasing computational cost and pipeline complexity. 
Another line of work adopts training-free strategies to promote exploration at inference time, including diverse prompting~\citep{dipper}, weight interpolation~\citep{wiseft}, entropy-aware temperature adjustment~\citep{28rule}, and latent-posterior manipulation~\citep{led}.
However, these methods mainly encourage broader exploration without explicitly guiding it toward continuations that make meaningful progress.
As illustrated in Fig.~\ref{fig:intro_motivation}, unguided exploration may still produce redundant or unproductive candidates, requiring more rollouts to reach a correct solution, i.e., exhibiting low \emph{exploration efficiency}. 
Higher exploration efficiency instead indicates a greater chance of covering at least one correct solution with a smaller rollout budget.

To improve exploration efficiency, we propose \emph{\underline{S}tate-conditioned \underline{P}rogress-guided \underline{S}teering} (SPS), a training-free latent steering framework for reasoning models. 
Our key intuition is that high-uncertainty reasoning transitions often correspond to critical forks that can lead to different reasoning paths~\citep{zheng2025first,dong2026agentic}. 
At such transitions, SPS applies progress-guided steering to guide the next reasoning step toward meaningful progress. 
To support this online steering, SPS constructs a state-conditioned direction bank offline from a small calibration dataset. 
The bank contains multiple progress-guided steering vectors for different prefix-state regions, derived from contrasting progress-making and non-progress-making continuations. 
During online inference, SPS retrieves a suitable steering vector based on the current prefix state and applies it at high-uncertainty transitions. 
SPS operates only during generation and is independent of the verifier or selection rule applied afterward.
Extensive experiments across multiple model scales and nine benchmarks consistently demonstrate the effectiveness of SPS. 

To summarize, our contributions are as follows:

\vspace{0.3em}

\quad \textbf{1)} We propose SPS, a training-free latent steering framework that explicitly guides reasoning exploration toward meaningful progress via state-conditioned steering, improving exploration efficiency under limited rollout budgets. SPS requires no additional training and is independent of the downstream verifier or selection rule.

\vspace{0.3em}

\quad \textbf{2)} Extensive experiments across multiple model scales and nine benchmarks show that SPS outperforms all baseline methods. Specifically, SPS consistently improves both Pass@4 and Pass@1, achieving a 4.39\% relative gain in average Pass@4 over strong baseline LED on Qwen3-4B.

\vspace{0.3em}

\quad \textbf{3)} Further analyses and ablations validate the key design choices of SPS and reveal that state awareness and appropriate intervention depth are key to effective steering. 
Our broader experiments also offer additional insights for future research.

\section{Method}
\label{sec:method}
In this section, we present SPS, a state-conditioned, progress-guided latent steering framework for efficient reasoning exploration. We first formalize the problem setting and provide an overview of SPS (\S\ref{sec:sps_overview}). We then describe how SPS constructs prefix-specific progress-guided latent directions (\S\ref{sec:direction_construction}) and organizes them into a state-conditioned direction bank (\S\ref{sec:bank_construction}). Finally, we introduce the online state-conditioned steering procedure (\S\ref{sec:online_steering}).

\subsection{Problem Setup and Method Overview}
\label{sec:sps_overview}
\begin{figure*}[!t]
    \centering \centerline{\includegraphics[width=2\columnwidth]{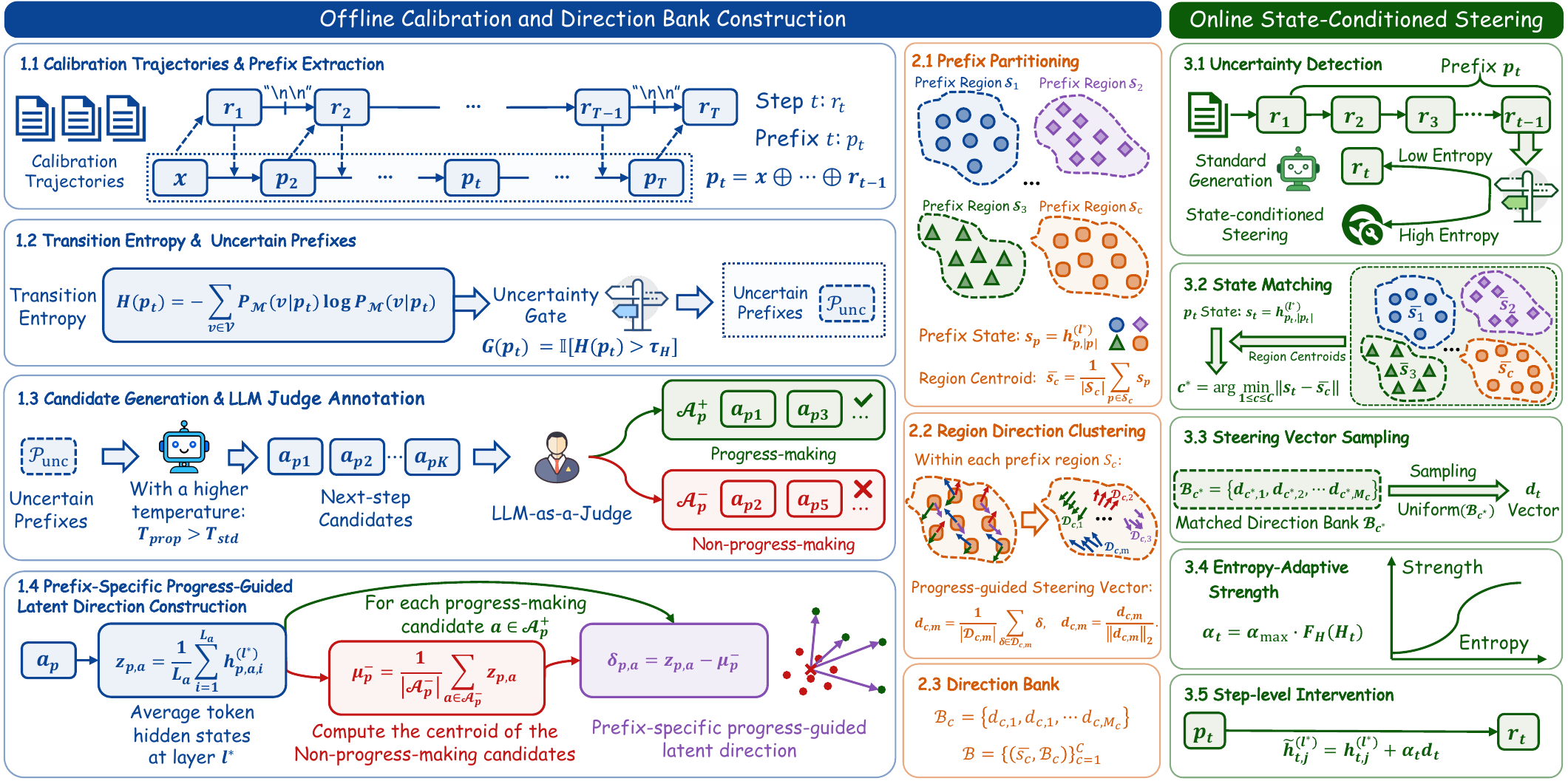}}
    \vspace{-0.1cm}
    \caption{The left panel shows the construction of prefix-specific progress-guided latent directions: SPS collects calibration trajectories, identifies high-uncertainty transitions, samples candidate next steps, and contrasts progress-making and non-progress-making candidates to construct prefix-specific directions (\S\ref{sec:direction_construction}). The middle panel shows the construction of the state-conditioned Direction Bank, where prefix states and their associated directions are clustered to form multiple steering vectors for each region (\S\ref{sec:bank_construction}). The right panel shows online steering: at each high-uncertainty transition, SPS matches the current prefix state to a region and applies a sampled steering vector to guide the next reasoning step (\S\ref{sec:online_steering}). See Appendix~\ref{app:sps_algo} for the algorithmic description of SPS.}
    \label{fig:sps_framework}
\end{figure*}

\paragraph{Problem Setup.}
Let $\mathcal{M}$ denote a reasoning model and let $x$ denote an input problem. Given a rollout budget $N$, the model generates a set of independent reasoning trajectories
\begin{equation}
\resizebox{\linewidth}{!}{%
$\displaystyle
    \mathcal{Y}_N(x)
    =
    \left\{
        y^{(1)}, y^{(2)}, \ldots, y^{(N)}
    \right\}, \quad
    y^{(n)} \sim \mathcal{M}(\cdot \mid x).$%
}
\end{equation}
In a Best-of-$N$ pipeline, a verifier or other selector chooses one candidate from $\mathcal{Y}_N(x)$ as the final output. Our goal is to improve the \emph{exploration efficiency} of $\mathcal{M}$, i.e., to increase the chance that $\mathcal{Y}_N(x)$ contains at least one correct reasoning trajectory with a smaller rollout budget $N$.

\paragraph{Method Overview.}
Prior work has shown that high-uncertainty transitions often correspond to critical forks that can lead to different reasoning paths~\citep{zheng2025first,dong2026agentic}. 
Across independently sampled trajectories, unguided generation at such transitions may repeatedly enter unproductive continuations, making it less likely to reach a correct solution within a limited rollout budget. 
To address this, SPS aims to steer model generation toward meaningful progress at such transitions during online inference. 
SPS operates only during generation and is independent of the verifier or selection rule applied afterward.

To support this online guidance, SPS constructs a state-conditioned direction bank offline. 
It first generates reasoning trajectories for a small calibration dataset and identifies high-uncertainty transitions in these trajectories. 
At each such transition, the same reasoning model generates several candidate continuations using a higher proposal temperature. 
An LLM judge labels these candidates as progress-making or non-progress-making. 
SPS contrasts the representations of these two groups to derive prefix-specific progress-guided latent directions (\S\ref{sec:direction_construction}). These directions are then organized into the state-conditioned direction bank (\S\ref{sec:bank_construction}).

During online inference (\S\ref{sec:online_steering}), SPS intervenes at high-uncertainty transitions. At each such transition, it retrieves a progress-guided steering vector from the direction bank based on current prefix state and applies the vector while generating next reasoning step. 
See Fig.~\ref{fig:sps_framework} for the overview of SPS.

\subsection{Prefix-Specific Progress-Guided Latent Direction Construction}
\label{sec:direction_construction}
\paragraph{High-Uncertainty Transition Identification and Candidate Generation.}
Let $\mathcal{D}_{\mathrm{cal}}=\{x_i\}_{i=1}^{M}$ denote the calibration dataset. For each problem $x_i\in\mathcal{D}_{\mathrm{cal}}$, we generate a reasoning trajectory $y_i$ with $\mathcal{M}$ using the standard inference temperature $T_{\mathrm{std}}$. 
We split $y_i$ into reasoning steps with delimiter \texttt{\textbackslash n\textbackslash n} and define the context before each reasoning step $r_{i,t}$ as its prefix:

\begin{equation}
\label{eq:reasoning_prefix}
\begin{aligned}
    y_i
    &=
    \left(
        r_{i,1}, r_{i,2}, \ldots, r_{i,T_i}
    \right), \\
    p_{i,t}
    &=
    x_i
    \oplus
    r_{i,1}
    \oplus
    \cdots
    \oplus
    r_{i,t-1}.
\end{aligned}
\end{equation}

\noindent where $r_{i,t}$ denotes the $t$-th reasoning step and $\oplus$ denotes text concatenation. We refer to the boundary $p_{i,t}\rightarrow r_{i,t}$ as a reasoning transition.

To identify high-uncertainty transitions, we first compute the \emph{transition entropy}, defined as the entropy of the next-token distribution used to generate the first token of $r_{i,t}$, and then utilize it to define the following entropy-based \emph{Uncertainty Gate}:
\begin{equation}
\label{eq:uncertainty_gate}
\begin{aligned}
    H(p_{i,t})
    =
    -\textstyle\sum_{v\in\mathcal{V}}
    &P_{\mathcal{M}}(v\mid p_{i,t})
    \log P_{\mathcal{M}}(v\mid p_{i,t}), \\
    G(p_{i,t})
    &=
    \mathbb{I}\!\left[
        H(p_{i,t})>\tau_H
    \right].
\end{aligned}
\end{equation}

\noindent where $\tau_H=Q_H(q_H)$, $Q_H(\cdot)$ is the empirical quantile function~\citep{hyndman1996sample} of $H(p_{i,t})$ over all calibration transitions, and $q_H\in(0,1)$ is the quantile level.
The prefixes that pass the gate form the following set of high-uncertainty prefixes:
\begin{equation}
\label{eq:uncertain_prefix_set}
\resizebox{\linewidth}{!}{%
$\displaystyle
    \mathcal{P}_{\mathrm{unc}}
    =
    \left\{
        p_{i,t}
        \;\middle|\;
        x_i\in\mathcal{D}_{\mathrm{cal}},\;
        1\leq t\leq T_i,\;
        G(p_{i,t})=1
    \right\}.
$%
}
\end{equation}

For each prefix $p\in\mathcal{P}_{\mathrm{unc}}$, we use the same model $\mathcal{M}$ with a higher proposal temperature $T_{\mathrm{prop}}>T_{\mathrm{std}}$ to generate $K$ candidate continuations:
\begin{equation}
    \mathcal{A}_{p}
    =
    \left\{
        a_{p,1}, a_{p,2}, \ldots, a_{p,K}
    \right\}.
    \label{eq:candidate_step_set}
\end{equation}
Generation stops at the next \texttt{\textbackslash n\textbackslash n} delimiter, so each candidate continuation contains one step. See \S\ref{para:implementation_details} for the choices of $\mathcal{D}_{\mathrm{cal}}$, $T_{\mathrm{std}}$, $\tau_H$, $T_{\mathrm{prop}}$, and $K$.

\paragraph{Progress-Based Candidate Annotation.}
For each prefix $p$ and candidate $a\in\mathcal{A}_p$, an LLM judge evaluates whether the candidate makes meaningful progress from the current reasoning context. The judge assigns either a positive label, \emph{progress-making}, or a negative label, \emph{non-progress-making}. We define the corresponding candidate sets as
\begin{equation}
\label{eq:candidate_groups}
\resizebox{0.93\linewidth}{!}{%
$\displaystyle
\begin{aligned}
    \mathcal{A}_p^{+}
    &=
    \left\{
        a\in\mathcal{A}_p
        \mid
        a\text{ is progress-making}
    \right\}, \\
    \mathcal{A}_p^{-}
    &=
    \left\{
        a\in\mathcal{A}_p
        \mid
        a\text{ is non-progress-making}
    \right\}.
\end{aligned}
$%
}
\end{equation}
Only prefixes for which both sets are non-empty are retained for latent direction construction.
The annotation prompt and additional details are provided in Appendix~\ref{app:llm_judge_annotation}.

\paragraph{Prefix-Specific Latent Direction Construction.}
Let $\ell^{\star}$ denote the intervention layer. We automatically select $\ell^{\star}$ based on the linear separability of progress-making and non-progress-making hidden representations across layers; details are provided in Appendix~\ref{app:layer_selection}. 
For each candidate continuation $a=(w_1,\ldots,w_{L_a})$, we average its token hidden states at layer $\ell^{\star}$ to obtain $z_{p,a}$, and then compute the centroid $\mu_p^{-}$ over the representations of non-progress-making candidates for each prefix $p$:
\begin{equation}
\label{eq:candidate_representation}
\resizebox{\linewidth}{!}{%
$\displaystyle
    z_{p,a}
    =
    \frac{1}{L_a}
    \sum_{j=1}^{L_a}
    h_{p,a,j}^{(\ell^{\star})},
    \quad
    \mu_p^{-}
    =
    \frac{1}{|\mathcal{A}_p^{-}|}
    \sum_{a\in\mathcal{A}_p^{-}}
    z_{p,a}.
$%
}
\end{equation}
For each progress-making candidate $a\in\mathcal{A}_p^{+}$, we derive a latent direction by contrasting its representation with the non-progress-making centroid:
\begin{equation}
    \delta_{p,a}
    =
    z_{p,a}-\mu_p^{-}.
    \label{eq:prefix_specific_direction}
\end{equation}
Here, $\delta_{p,a}$ points from the non-progress-making centroid toward the progress-making candidate. Since $z_{p,a}$ and $\mu_p^{-}$ are both computed from candidates generated from the same prefix $p$, we call $\delta_{p,a}$ a \emph{prefix-specific progress-guided latent direction}. Since a prefix $p$ may have multiple progress-making candidates, it may be associated with multiple directions $\{\delta_{p,a}\mid a\in\mathcal{A}_p^{+}\}$.

\subsection{State-Conditioned Direction Bank Construction}
\label{sec:bank_construction}
The prefix-specific directions constructed above are tied to their individual source prefixes, while online prefixes may differ from those seen offline. 
To make these directions applicable beyond their source prefixes during online inference, SPS first groups offline prefixes with similar latent states into several regions. 
An online prefix can then obtain guidance from its best-matched region.
Within each region, the associated prefix-specific directions may further reflect several common progress patterns. SPS therefore clusters these directions and uses each cluster centroid as a progress-guided steering vector.
We describe these two steps below.

\paragraph{Prefix Partitioning.}
To group similar prefixes, we denote each selected prefix $p\in\mathcal{P}_{\mathrm{unc}}$ by the hidden state of its last token at layer $\ell^{\star}$, $s_p=h_{p,|p|}^{(\ell^{\star})}$, which we refer to as \emph{prefix state}. 
We partition these prefixes into $C$ regions by clustering their prefix states,
$\{\mathcal{S}_1,\ldots,\mathcal{S}_C\}$,
where $\mathcal{S}_c$ contains the prefixes assigned to region $c$.
Its centroid is
\begin{equation}
    \bar{s}_c
    =
    \frac{1}{|\mathcal{S}_c|}
    \sum_{p\in\mathcal{S}_c}
    s_p.
    \label{eq:state_centroid}
\end{equation}

\paragraph{Within-Region Direction Clustering.}
For each region $\mathcal{S}_c$, we collect the prefix-specific directions $\delta_{p,a}$ associated with prefixes $p\in\mathcal{S}_c$:
\begin{equation}
    \Delta_c
    =
    \left\{
        \delta_{p,a}
        \;\middle|\;
        p\in\mathcal{S}_c,\;
        a\in\mathcal{A}_p^{+}
    \right\}.
    \label{eq:state_prefix_directions}
\end{equation}
These directions may reflect different ways of making progress. Averaging them into a single steering vector would remove these differences. We therefore cluster the directions in $\Delta_c$ into $M_c$ groups $\{\mathcal{D}_{c,1},\ldots,\mathcal{D}_{c,M_c}\}$. For each group $\mathcal{D}_{c,m}$, we compute its centroid and normalize it:
\begin{equation}
    \bar{d}_{c,m}
    =
    \frac{1}{|\mathcal{D}_{c,m}|}
    \sum_{\delta\in\mathcal{D}_{c,m}}
    \delta,
    \quad
    d_{c,m}
    =
    \frac{\bar{d}_{c,m}}
    {\|\bar{d}_{c,m}\|_2}.
    \label{eq:steering_vector}
\end{equation}
Each normalized centroid $d_{c,m}$ serves as a \emph{progress-guided steering vector}. The $M_c$ steering vectors form the direction bank for state region $c$, and the complete state-conditioned \emph{Direction Bank} is
\begin{equation}
\label{eq:complete_direction_bank}
\resizebox{\linewidth}{!}{%
$\displaystyle
    \mathcal{B}_c
    =
    \left\{
        d_{c,1}, d_{c,2}, \ldots, d_{c,M_c}
    \right\},
    \quad
    \mathcal{B}
    =
    \left\{
        \left(
            \bar{s}_c,\mathcal{B}_c
        \right)
    \right\}_{c=1}^{C}.
$%
}
\end{equation}
Thus, $\mathcal{B}$ contains $C$ state regions, each represented by its centroid $\bar{s}_c$ and paired with a direction bank $\mathcal{B}_c$ containing $M_c$ progress-guided steering vectors. 
See Appendix~\ref{app:clustering_details} for clustering details.

\subsection{Online State-Conditioned Steering}
\label{sec:online_steering}
\paragraph{State Matching and Steering Vector Sampling.}
Let $p_t$ denote the current online prefix before generating step $r_t$, and $p_t\rightarrow r_t$ denote the current reasoning transition. Let $H_t=H(p_t)$ denote the entropy at this transition. SPS applies the same uncertainty gate $G$ defined in Eq.~\ref{eq:uncertainty_gate}. If $G(p_t)=0$, SPS continues standard generation. If $G(p_t)=1$, SPS extracts the current prefix state $s_t=h_{p_t,|p_t|}^{(\ell^{\star})}$.

To retrieve a suitable steering vector, SPS performs \emph{State Matching} by matching $s_t$ to the region whose centroid is closest to it:
\begin{equation}
    c^{\star}
    =
    \arg\min_{1\leq c\leq C}
    \left\|
        s_t-\bar{s}_c
    \right\|_2.
    \label{eq:state_matching}
\end{equation}
This identifies the matched region $c^{\star}$ and its direction bank $\mathcal{B}_{c^{\star}}$.
To encourage diversity across rollouts, SPS performs \emph{Steering Vector Sampling} by randomly sampling one progress-guided steering vector $d_t$ from $\mathcal{B}_{c^{\star}}$:
\begin{equation}
    d_t
    \sim
    \mathrm{Uniform}\!\left(
        \mathcal{B}_{c^{\star}}
    \right).
    \label{eq:steering_vector_sampling}
\end{equation}
Since $\mathcal{B}_{c^{\star}}$ contains multiple progress-guided steering vectors for the matched region, sampling different vectors across rollouts allows the model to follow diverse progress-making directions.

\paragraph{Entropy-Adaptive Strength and Intervention.}
To apply stronger guidance when the model is more uncertain, SPS uses \emph{Entropy-Adaptive Strength}. 
Let $F_H$ denote the empirical cumulative distribution function (CDF) of transition entropy values $H(p_{i,t})$ over all reasoning transitions in the calibration trajectories. 
we set its steering strength according to the percentile rank of $H_t$ under this calibration distribution:
\begin{equation}
    \alpha_t
    =
    F_H(H_t),
    \label{eq:entropy_adaptive_strength}
\end{equation}
where $\alpha_t \in [0,1]$ is the resulting entropy percentile.
For example, if $H_t$ lies at the 80th percentile, then $\alpha_t=0.80$. A higher transition uncertainty therefore results in a larger steering strength.
Starting from the last token of $p_t$, SPS intervenes at each autoregressive generation position until $r_t$ ends:
\begin{equation}
    \widetilde{h}_{t,j}^{(\ell^{\star})}
    =
    h_{t,j}^{(\ell^{\star})}
    +
    \alpha_t d_t.
    \label{eq:step_intervention}
\end{equation}
The same $d_t$ and $\alpha_t$ are used throughout the generation of $r_t$. At the next reasoning transition $p_{t+1}\rightarrow r_{t+1}$, SPS evaluates $G(p_{t+1})$ again.

\section{Experiments}
\label{sec:exp}

\subsection{Experimental Setup}
\label{sec:exp_setting}
\paragraph{Benchmarks.}
We primarily evaluate SPS on six mathematical reasoning benchmarks: MATH-500~\citep{math500}, AIME 2024~\citep{aime24}, AIME 2025~\citep{aime25}, Minerva-Math~\citep{Minerva}, OlympiadBench~\citep{olympiadbench}, and HMMT 2025~\citep{hmmt25}. 
To further assess its generalization across domains, we also evaluate SPS on GPQA-Diamond~\citep{gpqa} for science, LiveCodeBench~\citep{livecodebench} for coding, and StrategyQA~\citep{strategyqa} for commonsense reasoning.
See Appendix~\ref{app:benchmark_details} for more benchmark details.

\paragraph{Models and Metrics.} We conduct experiments on Qwen3 models~\citep{qwen3} ranging from 1.7B to 14B. 
We report Pass@$K$ to assess the model's exploration capability, measuring whether a problem is solved correctly at least once within $K$ attempts, and Pass@1 (averaged over $K$ runs) to measure overall model performance. 
Prior work commonly uses $K=16$ or $32$~\citep{guo2025deepseek,qwen3,liang2025sws}; we instead use $K=4$ as our primary setting to focus on exploration with a limited rollout budget.

\paragraph{Baselines.} We compare SPS with five baselines: (i) \emph{CoT}~\citep{cot}, which uses standard chain-of-thought reasoning; (ii) \emph{DIPPER}~\citep{dipper}, which promotes reasoning diversity by sampling from multiple reasoning prompts; (iii) \emph{WiSE-FT}~\citep{wiseft}, which interpolates model weights to recover reasoning diversity; (iv) \emph{ETT}~\citep{28rule}, which assigns a higher temperature to high-entropy positions; and (v) \emph{LED}~\citep{led}, which leverages higher-entropy intermediate-layer posteriors to restore exploration. See Appendix~\ref{app:baseline_details} for baseline details. 

\paragraph{Implementation Details.} \label{para:implementation_details} 
Unless otherwise specified, all methods are evaluated in thinking mode with the same standard sampling settings that are widely adopted in prior work~\citep{qwen3}: $T_{\mathrm{std}}=0.6$, top-$p=0.95$, top-$k=20$, and a maximum generation length of $32{,}768$ tokens. 
For candidate generation, we set $T_{\mathrm{prop}}=1.2$ and $K=32$. We construct $\mathcal{D}_{\mathrm{cal}}$ by randomly sampling 200 prompts from DAPO-Math-17K~\citep{dapo}; sensitivity to the calibration dataset size is studied in Appendix~\ref{app:calibration_dataset_size}. Inspired by the high-entropy token analysis of~\citet{28rule}, we set $q_H=0.8$ for the uncertainty gate. 
We use vLLM~\citep{vllm} for inference and conduct experiments on NVIDIA H20 GPUs, with efficiency analysis in Appendix~\ref{app:efficiency}.
Further implementation details are provided in Appendix~\ref{app:sps_framework_details}.

\subsection{Performance of SPS}
\label{sec:main_results}
\begin{table*}[!t]
    \centering
    \renewcommand{\arraystretch}{1.15}
    \fontsize{8}{9}\selectfont
    \setlength{\tabcolsep}{3pt}

    \resizebox{\textwidth}{!}{
    \begin{tabular}{l cc cc cc cc cc cc}
        \toprule[1.5pt]
        & \multicolumn{2}{c}{\textbf{MATH}}
        & \multicolumn{2}{c}{\textbf{Minerva}}
        & \multicolumn{2}{c}{\textbf{Olympiad}}
        & \multicolumn{2}{c}{\textbf{AIME24}}
        & \multicolumn{2}{c}{\textbf{AIME25}}
        & \multicolumn{2}{c}{\textbf{HMMT25}} \\
        \cmidrule(lr){2-3}
        \cmidrule(lr){4-5}
        \cmidrule(lr){6-7}
        \cmidrule(lr){8-9}
        \cmidrule(lr){10-11}
        \cmidrule(lr){12-13}

        \textbf{Method}
        & Pass@1 & Pass@4
        & Pass@1 & Pass@4
        & Pass@1 & Pass@4
        & Pass@1 & Pass@4
        & Pass@1 & Pass@4
        & Pass@1 & Pass@4 \\
        \midrule

        \rowcolor{gray!20}
        \multicolumn{13}{c}{\textbf{Qwen3-1.7B}} \\
        \addlinespace[1pt]

        CoT
        & 89.35 & 95.00
        & 42.56 & 53.31
        & 64.11 & 74.81
        & 42.50 & 63.33
        & 33.33 & 46.67
        & 22.50 & 33.33 \\

        DIPPER
        & 89.45 & 95.20
        & 42.28 & 52.94
        & 64.22 & 75.11
        & 41.67 & 63.33
        & 31.67 & 46.67
        & 21.67 & 33.33 \\

        WiSE-FT
        & 89.20 & 94.80
        & 42.74 & 53.68
        & 63.93 & 74.52
        & 40.83 & 66.67
        & 30.00 & 43.33
        & 20.83 & 36.67 \\

        ETT
        & 89.70 & 95.60
        & 43.20 & 54.41
        & 64.52 & 75.56
        & 45.00 & 66.67
        & 32.50 & 53.33
        & 25.00 & 40.00 \\

        LED
        & 89.60 & 95.40
        & 43.57 & 55.15
        & 64.70 & 75.85
        & 45.83 & 70.00
        & 34.17 & 56.67
        & 24.17 & 36.67 \\

        \textbf{SPS (Ours)}
        & \textbf{90.15} & \textbf{96.40}
        & \textbf{44.21} & \textbf{56.62}
        & \textbf{66.04} & \textbf{78.22}
        & \textbf{49.17} & \textbf{76.67}
        & \textbf{40.83} & \textbf{63.33}
        & \textbf{28.33} & \textbf{46.67} \\

        \midrule

        \rowcolor{gray!20}
        \multicolumn{13}{c}{\textbf{Qwen3-4B}} \\
        \addlinespace[1pt]

        CoT
        & 94.45 & 97.20
        & 48.53 & 53.68
        & 75.93 & 83.41
        & 70.83 & 76.67
        & 63.33 & 70.00
        & 42.50 & 53.33 \\

        DIPPER
        & 94.70 & 97.60
        & 48.71 & 54.04
        & 76.07 & 83.70
        & 70.00 & 76.67
        & 62.50 & 66.67
        & 40.83 & 50.00 \\

        WiSE-FT
        & 94.25 & 97.40
        & 47.98 & 52.94
        & 76.30 & 84.15
        & 68.33 & 80.00
        & 60.83 & 73.33
        & 40.00 & 56.67 \\

        ETT
        & 94.60 & 97.40
        & 49.54 & 55.51
        & 76.19 & 84.00
        & 72.50 & 80.00
        & 66.67 & 76.67
        & 45.00 & 60.00 \\

        LED
        & 94.80 & 97.80
        & 49.36 & 55.15
        & 76.52 & 84.44
        & 74.17 & 83.33
        & 65.83 & 73.33
        & 45.83 & 56.67 \\

        \textbf{SPS (Ours)}
        & \textbf{95.15} & \textbf{98.40}
        & \textbf{50.37} & \textbf{57.35}
        & \textbf{77.56} & \textbf{86.37}
        & \textbf{76.67} & \textbf{86.67}
        & \textbf{70.00} & \textbf{83.33}
        & \textbf{48.33} & \textbf{66.67} \\

        \bottomrule[1.5pt]
    \end{tabular}
    }

    \vspace{-0.15cm}
    \caption{
    Results of SPS and baseline methods on Qwen3-1.7B and Qwen3-4B across six mathematical reasoning benchmarks.
    See Appendix~\ref{app:baseline_details} for the details of baseline methods.
    }
    \label{tab:main_results_small}
    \vspace{-0.2cm}
\end{table*}

Table~\ref{tab:main_results_small} reports the detailed performance of SPS and all baseline methods on Qwen3-1.7B and Qwen3-4B across six mathematical reasoning benchmarks.
The results show that SPS consistently outperforms all baselines in both Pass@1 and Pass@4.
Notably, the improvements are more pronounced on Pass@4, indicating that SPS is particularly effective at increasing the chance of finding a correct solution under a limited rollout budget.
For example, on Qwen3-1.7B, SPS improves Pass@4 over the strong baseline LED by 2.67\%, 3.12\%, and 9.53\% on Minerva, Olympiad, and AIME24, respectively.
SPS also consistently improves Pass@1, showing that progress-guided steering not only enhances exploration across multiple rollouts, but also improves the quality of individual generations.
We also provide qualitative case studies of SPS at high-uncertainty reasoning transitions in Appendix~\ref{app:case_study}.
These observations indicate that:
\begin{SummaryBox}
\textit{Explicitly guiding exploration toward meaningful progress enables more effective reasoning, especially under small rollout budgets.}
\end{SummaryBox}

\subsection{Further Analysis}
\label{sec:further_analysis}
\paragraph{SPS Across Different Rollout Budgets.}
\begin{figure}[!h]
    \centering \centerline{\includegraphics[width=\columnwidth]{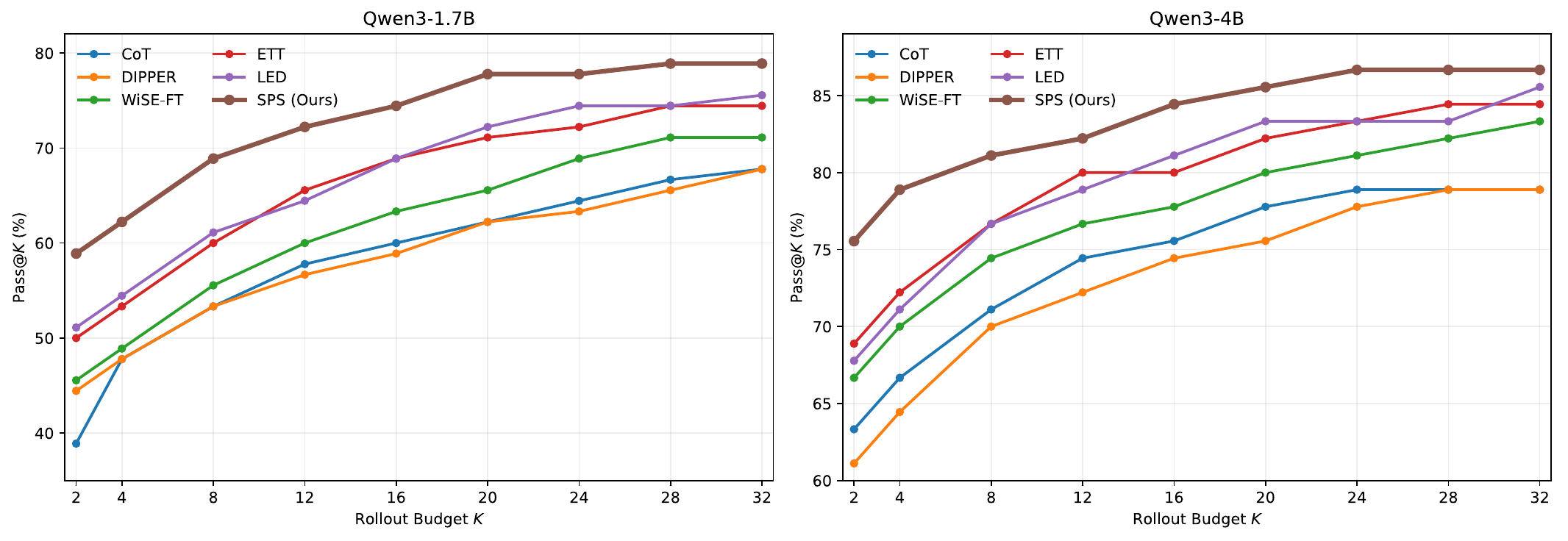}}
    \vspace{-0.2cm}
    \caption{Average Pass@$K$ of SPS and baselines across AIME 2024, AIME 2025, and HMMT 2025 under different rollout budgets for Qwen3-1.7B and Qwen3-4B.}
    \vspace{-0.2cm}
    \label{fig:rollout_budget_scaling}
\end{figure}

We report the average Pass@$K$ of SPS and baseline methods for Qwen3-1.7B and Qwen3-4B across different rollout budgets in Fig.~\ref{fig:rollout_budget_scaling}.
As shown in Fig.~\ref{fig:rollout_budget_scaling}, SPS consistently outperforms all baseline methods across different rollout budgets.
The advantage is particularly pronounced under smaller rollout budgets, especially from $K=2$ to $K=16$, demonstrating that SPS is more effective at discovering correct solutions when only a limited number of rollouts are available.
For instance, on Qwen3-4B, SPS achieves improvements of 4.34\% and 4.11\% over the competitive baseline LED at $K=8$ and $K=16$, respectively.
Importantly, SPS establishes its advantage at small rollout budgets and maintains the strongest performance throughout the evaluated range, showing that its benefit is not limited to a particular rollout budget.
Based on these observations, we conclude:
\begin{SummaryBox}
\textit{SPS achieves higher exploration efficiency by discovering correct solutions with fewer rollouts, yielding particularly strong performance under limited rollout budgets.}
\end{SummaryBox}

\paragraph{SPS on Larger Model Scale.}
\begin{table}[!t]
    \centering

    \renewcommand{\arraystretch}{1.2}

    \fontsize{9.5}{8}\selectfont

    \setlength{\tabcolsep}{6pt}

    \begin{tabular}{l cccc}
        \toprule[1.5pt]
        & \multicolumn{2}{c}{\textbf{Qwen3-8B}}
        & \multicolumn{2}{c}{\textbf{Qwen3-14B}} \\
        \cmidrule(lr){2-3} \cmidrule(lr){4-5}
        & Pass@1  & Pass@4 
        & Pass@1  & Pass@4  \\
        \midrule

        CoT
        & 67.59 & 75.15
        & 70.58 & 78.57 \\

        \addlinespace[1pt]

        DIPPER
        & 67.39 & 75.34
        & 70.83 & 78.28 \\

        \addlinespace[1pt]

        WiSE-FT
        & 66.62 & 75.78
        & 69.45 & 78.02 \\

        \addlinespace[1pt]

        ETT
        & 69.06 & 77.75
        & 72.14 & 80.75 \\

        \addlinespace[1pt]

        LED
        & 69.68 & 78.97
        & 72.53 & 81.31 \\

        \addlinespace[1pt]

        SPS (Ours)
        & \textbf{71.03} & \textbf{81.90}
        & \textbf{74.56} & \textbf{84.38} \\

        \bottomrule[1.5pt]
    \end{tabular}

    \vspace{-0.1cm}
    \caption{
    Average results of SPS and baselines across six mathematical reasoning benchmarks on Qwen3-8B and Qwen3-14B.
    See Table~\ref{tab:main_results_large} for detailed results.
    }
    \vspace{-0.4cm}
    \label{tab:larger_scale_results}
\end{table}
Having verified the effectiveness of SPS on Qwen3-1.7B and Qwen3-4B, we further evaluate its scalability to larger reasoning models.
As shown in Table~\ref{tab:larger_scale_results}, SPS consistently achieves the best average Pass@1 and Pass@4 across six benchmarks on both Qwen3-8B and Qwen3-14B.
On Qwen3-8B, SPS improves the average Pass@1 and Pass@4 over LED by 1.94\% and 3.71\%, respectively.
Similar improvements are observed on Qwen3-14B, where SPS surpasses LED by 2.79\% in average Pass@1 and 3.78\% in average Pass@4.
These results show that SPS remains effective as model scale increases, with consistent improvements in both Pass@1 and Pass@4.

\paragraph{Cross-Domain Generalization.}

\begin{figure}[!t]
    \centering
    \includegraphics[width=\columnwidth]{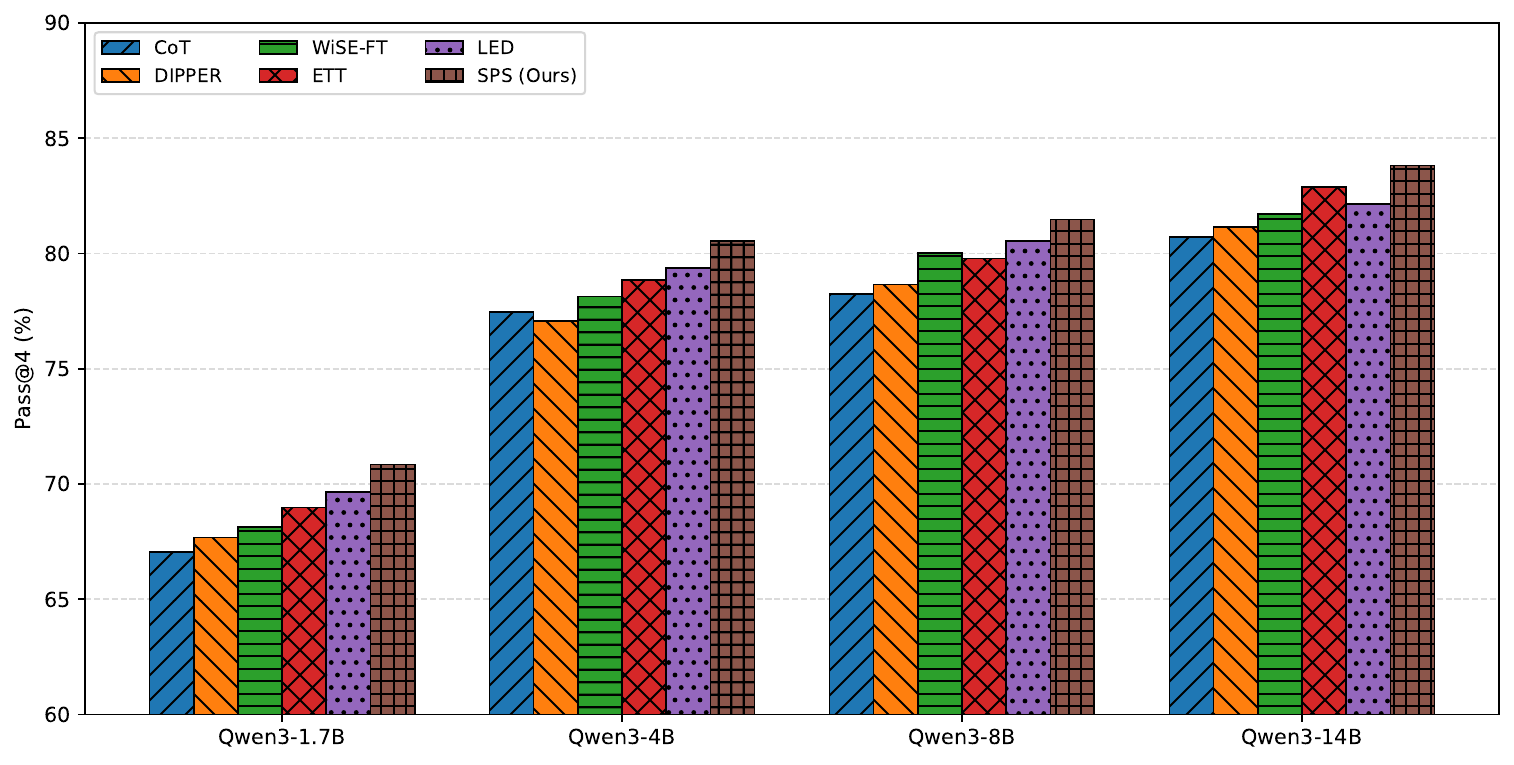}
    \vspace{-0.4cm}
    \caption{
    Average Pass@4 of SPS and baseline methods across GPQA-Diamond, StrategyQA, and LiveCodeBench.
    See Table~\ref{tab:generalization_detailed} for the detailed results.
    }
    \vspace{-0.2cm}
    \label{fig:generalization}
\end{figure}

We further evaluate SPS on three cross-domain benchmarks to examine its generalization beyond mathematics.
As shown in Fig.~\ref{fig:generalization}, SPS achieves the highest average Pass@4 across GPQA-Diamond, StrategyQA, and LiveCodeBench across all evaluated models.
For example, compared with LED on Qwen3-1.7B, SPS achieves relative improvements of 2.01\% and 1.74\% in average Pass@1 and Pass@4, respectively.
Similarly, on Qwen3-14B, SPS surpasses ETT by 1.32\% in average Pass@1 and 1.12\% in average Pass@4.
These results show that:
\begin{SummaryBox}
\textit{Progress-guided steering generalizes effectively across domains, enabling more effective reasoning exploration beyond mathematics.}
\end{SummaryBox}

\begin{figure}[!h]
    \centering
    \centerline{
        \includegraphics[width=\columnwidth]{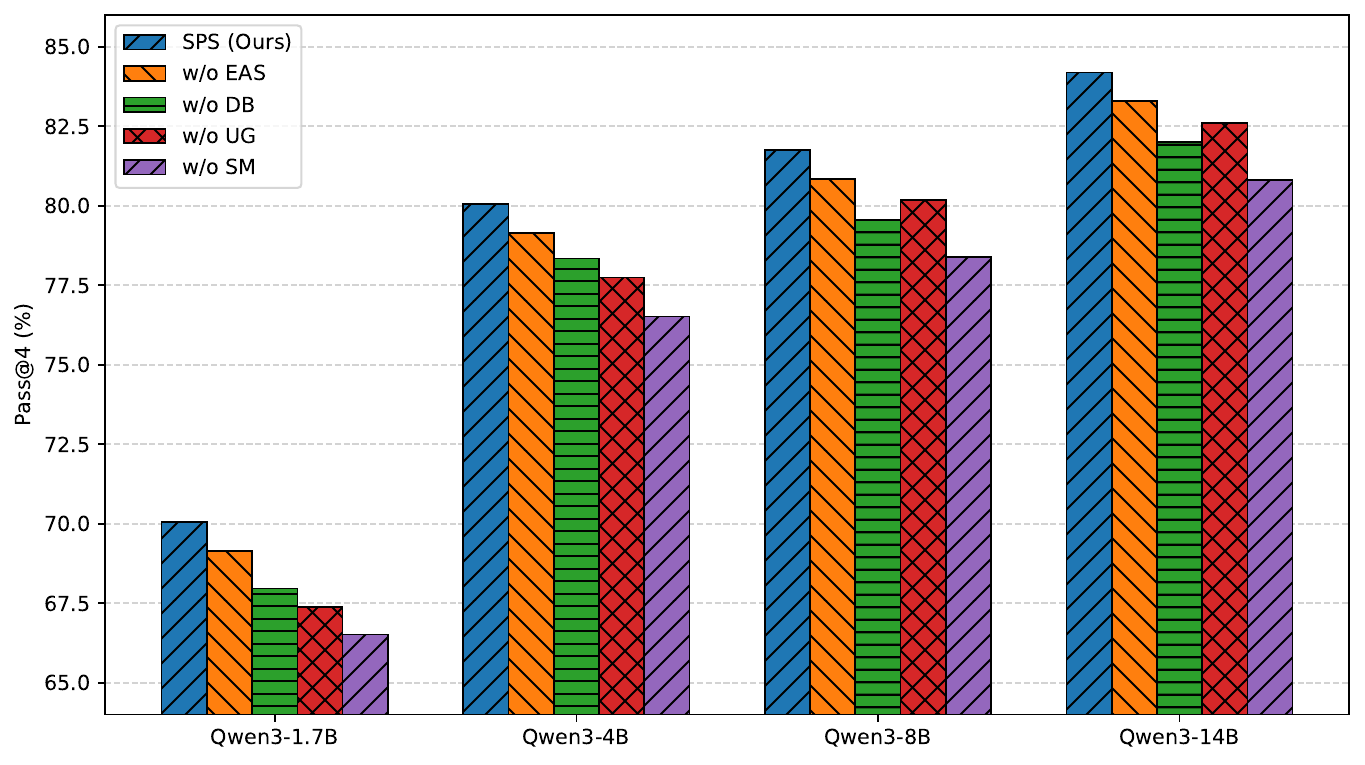}
    }
    \vspace{-0.3cm}
    \caption{
    Average Pass@4 of SPS and its ablation variants across nine benchmarks.
    ``w/o'' denotes removing the corresponding component from SPS.
    }
    \vspace{-0.2cm}
    \label{fig:ablation_pass4}
\end{figure}

\paragraph{Ablation Analysis.}
We conduct ablation experiments to examine the contributions of the \emph{Direction Bank (DB)}, \emph{State Matching (SM)}, \emph{Uncertainty Gate (UG)}, and \emph{Entropy-Adaptive Strength (EAS)} in SPS.
We report the average Pass@4 and Pass@1 across all nine benchmarks in Fig.~\ref{fig:ablation_pass4} and Fig.~\ref{fig:ablation_pass1}, respectively.
As shown in Fig.~\ref{fig:ablation_pass4}, removing any component consistently degrades performance, demonstrating that each component contributes to SPS.
Among all variants, removing \emph{State Matching} leads to the largest degradation, with a relative Pass@4 drop of up to 5.04\% on Qwen3-1.7B.
This highlights the importance of matching the current reasoning state with an appropriate direction bank for effective steering.
See Appendix~\ref{app:ablation_details} for more ablation details.
These observations indicate that:
\begin{SummaryBox}
\textit{All components contribute to SPS, while matching steering directions to the current reasoning state plays a more important role in effective state-conditioned steering.}
\end{SummaryBox}

\paragraph{Impact of the Steering Layer.}
\label{para:steering_layer}

\begin{figure}[!t]
    \centering
    \includegraphics[width=0.9\columnwidth]{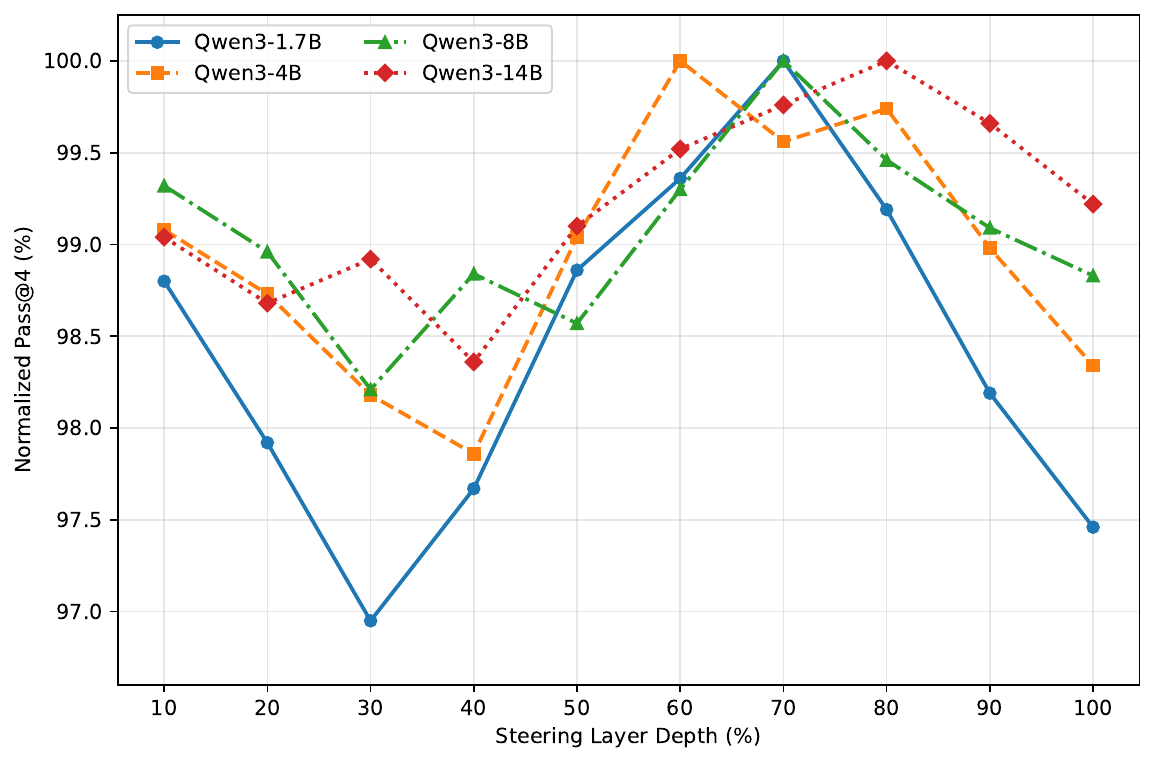}
    \vspace{-0.2cm}
    \caption{
Impact of the intervention layer on Pass@4, averaged over nine benchmarks and normalized by each model's best performance.
Layer depth denotes the layer index as a percentage of the total number of layers.
}
    \label{fig:steering_layer}
\end{figure}

We examine the impact of the intervention layer by applying SPS at different layers throughout the model.
As shown in Fig.~\ref{fig:steering_layer}, steering in mid-to-late layers generally yields the best performance across all model scales.
Specifically, the best performance occurs around $70\%$, $60\%$, $70\%$, and $80\%$ depth for Qwen3-1.7B, Qwen3-4B, Qwen3-8B, and Qwen3-14B, respectively.
Intervening too late becomes less effective, likely because fewer subsequent layers remain to propagate and integrate the steering signal.
This pattern is consistent with the linear probing analysis in Appendix~\ref{app:layer_selection}, where representations in similar layer regions exhibit the strongest linear separability.
These observations indicate that:
\begin{SummaryBox}
\textit{Steering is most effective in mid-to-late layers, where reasoning representations also exhibit the strongest linear separability.}
\end{SummaryBox}

\section{Related Work}
\label{sec:related_work}
\paragraph{Exploration Collapse in Reasoning Models.}
Modern reasoning models typically undergo supervised fine-tuning and reinforcement learning during post-training~\citep{qwen3,guo2025deepseek,ace_nemotron}. 
While these stages substantially improve Pass@1, they also induce \emph{exploration collapse}, where independent rollouts repeatedly follow similar reasoning paths, limiting the gains from Best-of-$N$ sampling at inference time~\citep{exploration_collapse_1,exploration_collapse_2,exploration_collapse_3}. 
To mitigate this problem, prior work modifies the training procedure to preserve reasoning diversity or explicitly encourage broader exploration~\citep{sft_diverse_1,sft_diverse_2,rl_diverse_1,rl_diverse_2,rl_diverse_4}. 
These approaches require additional training and often involve specialized data, loss functions, or reward designs, increasing computational cost and pipeline complexity.

To avoid retraining, another line of work promotes exploration directly at inference time. 
DIPPER~\citep{dipper} uses diverse reasoning prompts to induce different reasoning paths across rollouts. 
\citet{wiseft} propose WiSE-FT, which interpolates model weights to recover reasoning diversity.
\citet{28rule} identify high-entropy tokens as critical reasoning forks and assign higher temperatures to these positions, while \citet{led} introduce LED, which restores exploration using higher-entropy intermediate-layer posteriors. 
However, these methods mainly increase exploration without explicitly guiding it toward continuations that make meaningful progress. 
Under a limited rollout budget, many candidates may therefore remain redundant or unproductive, reducing the chance of finding a correct solution and thereby limiting exploration efficiency.

\paragraph{Activation Steering for Large Language Models.}
Activation steering typically constructs steering vectors from contrasting positive and negative examples and injects them into a model's latent states during inference to control generation. It has shown effectiveness across various applications~\citep{activation_refusal_1,activation_refusal_2,glore,zhang2026locate,rebalance}.

In safety, \citet{activation_refusal_1} identify distinct harmfulness and refusal directions for controlling safety-related behavior, while \citet{activation_refusal_2} show that refusal directions transfer across safety-aligned languages.
For persona control, \citet{activation_persona_1} construct role vectors from model activations to steer role-specific behaviors, while \citet{activation_persona_2} identify an Assistant Axis for controlling and stabilizing the model's default persona.
Activation steering has also been applied to reasoning. \citet{activation_reason_1} derive reasoning control vectors to improve reasoning performance; GLORE~\citep{glore} uses representation engineering to elicit general long chain-of-thought reasoning; and ReBalance~\citep{rebalance} constructs a steering vector from reasoning-mode prototypes and dynamically adjusts it to balance overthinking and underthinking.
Different from these works, SPS constructs progress-guided steering vectors to guide exploration toward meaningful reasoning progress, increasing the chance of finding a correct solution under a limited rollout budget.

\section{Conclusion}
\label{sec:conclusion}
In this work, we propose SPS, a training-free latent steering framework for improving exploration efficiency in reasoning models. 
Unlike existing training-free methods that primarily increase reasoning diversity without explicitly guiding the exploration process, SPS constructs progress-guided steering vectors from progress-making and non-progress-making continuations to steer reasoning toward meaningful progress. 
By organizing these vectors into a state-conditioned direction bank and applying them at high-uncertainty reasoning transitions, SPS provides state-aware guidance while preserving diversity across rollouts, enabling more effective exploration under a limited rollout budget. 
Extensive experiments across different model scales and reasoning domains demonstrate that SPS consistently improves the chance of discovering correct solutions with limited rollouts compared to strong training-free baselines, without requiring additional model training. 
Our further analysis also offers valuable insights for future research.

\section*{Limitations}
\label{sec:limitations}
While SPS demonstrates consistent effectiveness across different model scales and reasoning domains, our experiments mainly focus on the Qwen3 model family with model sizes up to 14B parameters. It remains unclear whether the same findings generalize to substantially larger models or other model families. 
Future work will evaluate SPS on a broader range of reasoning models and larger model scales. 
We also plan to extend the evaluation to extra reasoning scenarios, such as multilingual and long-context reasoning, to further study the generalizability of our proposed framework.

\section*{Ethics Statement}
This work studies training-free latent steering for improving the exploration efficiency of reasoning models. All experiments are conducted on publicly available models and benchmarks. The human validation in our annotation analysis only involves reviewing model-generated reasoning steps and does not collect sensitive or personally identifiable information. Our method does not introduce new safety-critical applications or user-facing data collection. Nevertheless, as SPS can improve the reasoning capability of language models, it may inherit the general dual-use risks associated with advanced language models. We encourage responsible use of the proposed method in accordance with applicable model licenses and deployment policies.

\bibliography{custom.bib}

\clearpage
\appendix
\section{Algorithmic Description of SPS}
\label{app:sps_algo}
Algorithm~\ref{alg:sps} summarizes the complete SPS framework, including offline Direction Bank construction and online state-conditioned steering.

\begin{algorithm}[!h]
\caption{State-conditioned Progress-guided Steering (SPS)}
\label{alg:sps}
% \small
% \scriptsize
\fontsize{8.5pt}{7pt}\selectfont
\begin{algorithmic}[1]

\Require Reasoning model $\mathcal{M}$; calibration dataset $\mathcal{D}_{\mathrm{cal}}$; standard temperature $T_{\mathrm{std}}$; proposal temperature $T_{\mathrm{prop}}$; number of candidate steps $K$; entropy quantile $q_H$; minimum steering strength $\alpha_{\min}$

\Ensure State-conditioned Direction Bank $\mathcal{B}$ and SPS-guided generation

\Statex
\Statex \textbf{Offline Stage: Direction Bank Construction}

\ForAll{$x_i \in \mathcal{D}_{\mathrm{cal}}$}
    \State Generate a calibration trajectory $y_i$ with $\mathcal{M}$ using $T_{\mathrm{std}}$
    \State Split $y_i$ into reasoning steps $\{r_{i,t}\}_{t=1}^{T_i}$ and construct prefixes $\{p_{i,t}\}_{t=1}^{T_i}$
    \State Compute transition entropy $H(p_{i,t})$ for each prefix $p_{i,t}$
\EndFor

\State Compute $\tau_H = Q_H(q_H)$ and obtain
$\mathcal{P}_{\mathrm{unc}}
=
\{p_{i,t} \mid H(p_{i,t})>\tau_H\}$

\ForAll{$p \in \mathcal{P}_{\mathrm{unc}}$}
    \State Generate $K$ candidate next steps $\mathcal{A}_p$ using $T_{\mathrm{prop}}$
    \State Annotate candidates as progress-making or non-progress-making using the LLM judge
    \State Form $\mathcal{A}_p^{+}$ and $\mathcal{A}_p^{-}$; discard $p$ if either set is empty
\EndFor

\State Select the intervention layer $\ell^{\star}$ based on linear separability

\ForAll{retained prefixes $p$}
    \State Compute candidate representations
    $
    z_{p,a}
    =
    \frac{1}{L_a}
    \sum_{j=1}^{L_a}
    h_{p,a,j}^{(\ell^{\star})}
    $
    \State Compute
    $
    \mu_p^{-}
    =
    \frac{1}{|\mathcal{A}_p^{-}|}
    \sum_{a\in\mathcal{A}_p^{-}}
    z_{p,a}
    $
    \ForAll{$a \in \mathcal{A}_p^{+}$}
        \State Construct
        $
        \delta_{p,a}
        =
        z_{p,a}-\mu_p^{-}
        $
    \EndFor
    \State Extract prefix state
    $
    s_p=h_{p,|p|}^{(\ell^{\star})}
    $
\EndFor

\State Partition prefixes into regions
$\{\mathcal{S}_1,\ldots,\mathcal{S}_C\}$
based on their prefix states

\For{$c=1,\ldots,C$}
    \State Compute the region centroid
    $
    \bar{s}_c
    =
    \frac{1}{|\mathcal{S}_c|}
    \sum_{p\in\mathcal{S}_c}s_p
    $
    \State Collect
    $
    \Delta_c
    =
    \{\delta_{p,a}
    \mid
    p\in\mathcal{S}_c,\,
    a\in\mathcal{A}_p^{+}\}
    $
    \State Cluster $\Delta_c$ into
    $\{\mathcal{D}_{c,1},\ldots,\mathcal{D}_{c,M_c}\}$
    \For{$m=1,\ldots,M_c$}
        \State Compute
        $
        \bar{d}_{c,m}
        =
        \frac{1}{|\mathcal{D}_{c,m}|}
        \sum_{\delta\in\mathcal{D}_{c,m}}\delta
        $
        \State Normalize
        $
        d_{c,m}
        =
        \bar{d}_{c,m}/\|\bar{d}_{c,m}\|_2
        $
    \EndFor
    \State Form
    $
    \mathcal{B}_c
    =
    \{d_{c,1},\ldots,d_{c,M_c}\}
    $
\EndFor

\State Construct
$
\mathcal{B}
=
\{(\bar{s}_c,\mathcal{B}_c)\}_{c=1}^{C}
$

\Statex
\Statex \textbf{Online Stage: State-Conditioned Steering}

\State Given an input problem $x$, initialize the current prefix $p_1=x$

\For{$t=1,2,\ldots$ until generation terminates}
    \State Compute $H_t=H(p_t)$

    \If{$H_t \leq \tau_H$}
        \State Generate reasoning step $r_t$ using standard generation
    \Else
        \State Extract
        $
        s_t=h_{p_t,|p_t|}^{(\ell^{\star})}
        $
        \State Match the nearest region:
        $
        c^{\star}
        =
        \arg\min_{1\leq c\leq C}
        \|s_t-\bar{s}_c\|_2
        $
        \State Sample
        $
        d_t
        \sim
        \mathrm{Uniform}(\mathcal{B}_{c^{\star}})
        $
        \State Set
        $
        \alpha_t
        =
        F_H(H_t)
        $
        \State Generate $r_t$ while applying
        $
        \widetilde{h}_{t,j}^{(\ell^{\star})}
        =
        h_{t,j}^{(\ell^{\star})}
        +
        \alpha_t d_t
        $
        from the last token of $p_t$ until $r_t$ ends
    \EndIf

    \State Update
    $
    p_{t+1}
    =
    p_t \oplus r_t
    $
\EndFor

\end{algorithmic}
\end{algorithm}

\section{Details of SPS Framework}
\label{app:sps_framework_details}

\subsection{Candidate Step Deduplication}
\label{app:candidate_deduplication}
To improve the efficiency of LLM-as-Judge annotation, we remove highly similar candidate steps before annotation. We perform deduplication in two stages, considering both token-level and semantic similarity.

We first measure the token-level similarity between candidate pairs using ROUGE-L F1~\citep{lin2004rouge}, which captures their sequence overlap based on the longest common subsequence. Candidate pairs with a ROUGE-L F1 score above a threshold $\tau_{\mathrm{lex}}$ are treated as highly redundant and filtered.

We then encode the remaining candidates using Qwen3-Embedding-8B~\citep{qwen3embedding} and compute the cosine similarity between their embeddings. Candidate pairs with cosine similarity above $\tau_{\mathrm{sem}}$ are further treated as semantically redundant and filtered.

To determine $\tau_{\mathrm{lex}}$ and $\tau_{\mathrm{sem}}$, we manually select a set of highly similar candidate pairs and compute both similarity scores for each pair. We set each threshold to the average corresponding similarity score over these selected pairs. This two-stage filtering removes both lexically repetitive and semantically redundant candidate steps before LLM-as-Judge annotation.

\subsection{LLM-as-Judge Candidate Annotation and Label Validation}
\label{app:llm_judge_annotation}
\paragraph{LLM-as-Judge Candidate Annotation.}
We use GPT-5.6~\citep{openai2026frontier} with medium thinking effort to annotate the generated candidate steps. Given the original problem, the current reasoning prefix, and a candidate step, the judge evaluates its validity and local progress before assigning a final label. The complete annotation prompt is shown below.

\begin{tcolorbox}[
    enhanced,
    breakable,
    colback=gray!5,
    colframe=black,
    colbacktitle=black,
    coltitle=white,
    title=\textbf{LLM-as-Judge Candidate Annotation Prompt},
    fonttitle=\bfseries,
    boxrule=0.8pt,
    arc=2pt,
    left=6pt,
    right=6pt,
    top=6pt,
    bottom=6pt
]
\small

\textbf{Problem}\\
\texttt{\{problem\}}

\vspace{0.6em}
\textbf{Reasoning Prefix}\\
\texttt{\{reasoning\_prefix\}}

\vspace{0.6em}
\textbf{Candidate Step}\\
\texttt{\{candidate\_step\}}

\vspace{0.8em}
\textbf{Evaluation Criteria}

\begin{enumerate}
    \item \textbf{Validity}
    \begin{enumerate}
        \item \texttt{valid}: The step is mathematically and logically sound.
        \item \texttt{minor\_issue}: The core reasoning is sound, but there is a small imprecision, omission, or presentation issue.
        \item \texttt{invalid}: The step contains a substantive mathematical or logical error, contradicts the prefix, is irrelevant, or is not interpretable.
    \end{enumerate}

    \item \textbf{Local Progress}
    \begin{enumerate}
        \item \texttt{substantial}: The step makes clear progress by deriving a useful result, resolving an uncertainty, introducing a productive strategy, correcting an error, or eliminating an important possibility.
        \item \texttt{moderate}: The step provides some relevant progress, but the advancement is limited or incomplete.
        \item \texttt{minimal}: The step is mostly repetitive, superficial, or does not materially advance the reasoning.
        \item \texttt{harmful}: The step moves the reasoning in an incorrect, contradictory, or clearly unproductive direction.
    \end{enumerate}

    \item \textbf{Final Label}
    \begin{enumerate}
        \item \texttt{positive}: The step has validity \texttt{valid} or \texttt{minor\_issue} and makes \texttt{substantial} local progress.
        \item \texttt{negative}: The step is \texttt{invalid}, or its progress is \texttt{harmful}, or it is clearly repetitive, irrelevant, or unproductive.
        \item \texttt{neutral}: The step does not clearly satisfy either the positive or negative definition, including steps with only moderate or minimal but non-harmful progress.
    \end{enumerate}
\end{enumerate}

\textbf{Justification Requirements}

Provide a concise, evidence-based justification of 2--4 sentences. Explain whether the candidate is valid and how it does or does not advance the reasoning. Refer to specific claims, operations, or conclusions in the candidate step. Do not provide a full solution to the problem or lengthy hidden reasoning.

\vspace{0.6em}
Return only the following JSON object:

\begin{Verbatim}[
    breaklines=true,
    breaksymbolleft={},
    breaksymbolright={}
]
{
  "validity": "valid | minor_issue | invalid",
  "local_progress": "substantial | moderate | minimal | harmful",
  "label": "positive | neutral | negative",
  "confidence": 0.0,
  "justification": "Concise evidence-based explanation."
}
\end{Verbatim}
\end{tcolorbox}

Candidates labeled \texttt{positive} and \texttt{negative} are treated as progress-making and non-progress-making, respectively, while \texttt{neutral} candidates are excluded from latent direction construction.

\paragraph{Human Validation of LLM Annotations.}
To assess the reliability of the LLM-as-Judge annotations, we randomly sample 200 annotated candidates, including 100 positive and 100 negative examples. Two computer science PhD annotators independently review the same 200 examples. For each annotation, they examine whether the generated justification is correct and reasonable, and whether it supports the assigned final label. Each annotator assigns one of three decisions: \texttt{accept}, \texttt{reject}, or \texttt{uncertain}.

We report the acceptance rate of each reviewer and the proportion of examples accepted by both reviewers in Table~\ref{tab:llm_judge_validation}.\nocite{su2026attention,2026xufengsurvey}

\begin{table}[h]
    \centering
    \setlength\tabcolsep{3pt}
    \fontsize{9.5}{11.5}\selectfont
    \begin{tabular}{lccc}
        \toprule[1.5pt]
        \textbf{Subset}
        & \textbf{PhD 1 Accept}
        & \textbf{PhD 2 Accept}
        & \textbf{Both Accept} \\
        \midrule
        Positive & 94.0\% & 91.0\% & 88.0\% \\
        Negative & 90.0\% & 88.0\% & 85.0\% \\
        \midrule
        Overall  & 92.0\% & 89.0\% & 86.5\% \\
        \bottomrule[1.5pt]
    \end{tabular}
    \caption{Human validation of LLM-as-Judge annotations. Each subset contains 100 examples. ``Both Accept'' denotes the proportion of annotations accepted by both annotators.}
    \label{tab:llm_judge_validation}
\end{table}

\subsection{Automatic Intervention Layer Selection via Linear Separability}
\label{app:layer_selection}
To automatically select the intervention layer, we use the linear separability of latent representations as a selection criterion.

\paragraph{Linear Probing Setup.}
We randomly sample $2{,}000$ examples with balanced progress-making and non-progress-making labels, and split them into $80\%$ training and $20\%$ test sets.
For each example, we extract token-level hidden states at candidate layer depths from $10\%$ to $100\%$ and average them over all tokens to obtain a step-level representation.
Specifically, for a reasoning step $s$ containing $|s|$ tokens, its representation at layer $\ell$ is
\begin{equation}
    h_s^{(\ell)}
    =
    \frac{1}{|s|}
    \sum_{t=1}^{|s|}
    h_{s,t}^{(\ell)}.
\end{equation}

Since hidden representations are high-dimensional, we first apply PCA~\citep{pca} before probing.
For each probed layer $\ell$, the PCA dimensionality is chosen such that the retained variance exceeds $90\%$.
Let $h_s^{(\ell)}$ denote the hidden representation of example $s$ at layer $\ell$, and let
\begin{equation}
    z_s^{(\ell)}
    =
    \mathrm{PCA}_{\ell}
    \left(
        h_s^{(\ell)}
    \right)
\end{equation}
denote its reduced representation.

We then fit a ridge-regression linear probe on the training set:
\begin{equation}
\label{eq:layer_probe}
\resizebox{\linewidth}{!}{%
$\displaystyle
    \min_{w^{(\ell)},\,b^{(\ell)}}
    \sum_{s\in\mathcal{D}_{\mathrm{train}}}
    \left(
        y_s
        -
        {w^{(\ell)}}^{\top}z_s^{(\ell)}
        -
        b^{(\ell)}
    \right)^2
    +
    \lambda
    \left\|
        w^{(\ell)}
    \right\|_2^2,
    $%
}
\end{equation}
where $y_s\in\{0,1\}$ denotes the binary label.
The probe score for a test example is
\begin{equation}
    \hat{y}_s^{(\ell)}
    =
    {w^{(\ell)}}^{\top}z_s^{(\ell)}
    +
    b^{(\ell)}.
\end{equation}

We evaluate linear separability using the area under the ROC curve (AUC) on the held-out test set.
A higher AUC indicates that positive and negative representations are more linearly separable at the corresponding layer.
Accordingly, we automatically select the intervention layer as
\begin{equation}
    \ell^{\star}
    =
    \arg\max_{\ell}
    \mathrm{AUC}_{\ell}.
    \label{eq:automatic_layer_selection}
\end{equation}

\begin{figure}[!t]
    \centering
    \includegraphics[width=\columnwidth]{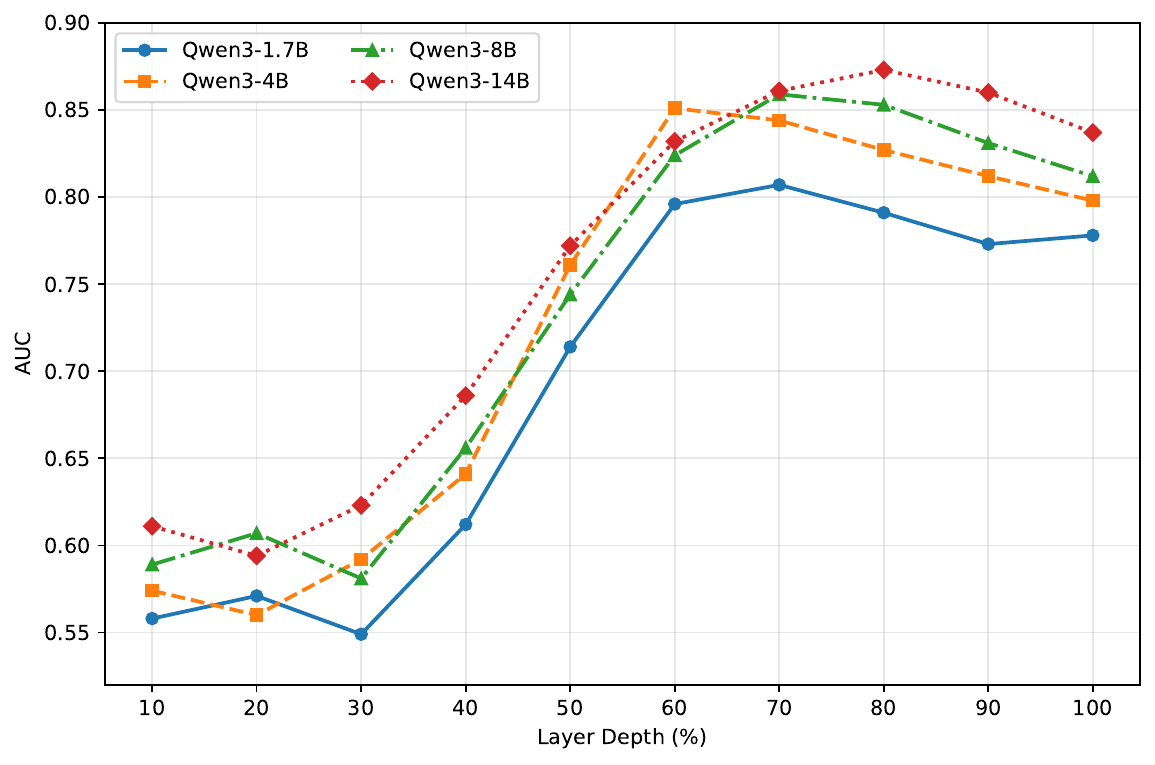}
    \vspace{-0.3cm}
    \caption{
    Probe AUC across normalized layer depths for four models.
    Higher AUC indicates stronger linear separability between progress-making and non-progress-making reasoning representations.
    }
    \label{fig:probing_auc_layers}
    \vspace{-0.35cm}
\end{figure}

\paragraph{Results.}
Fig.~\ref{fig:probing_auc_layers} shows that linear separability is relatively weak in early layers and increases substantially toward the middle and later layers.
The highest AUC is observed at approximately $70\%$, $60\%$, $70\%$, and $80\%$ layer depth for Qwen3-1.7B, Qwen3-4B, Qwen3-8B, and Qwen3-14B, respectively, corresponding to Layers $20$, $22$, $25$, and $32$.

Notably, these regions match the best intervention layers observed in Fig.~\ref{fig:steering_layer}.
This alignment suggests that stronger linear separability provides an effective proxy for identifying suitable intervention layers, avoiding exhaustive downstream evaluation across layers.
We therefore use the layer with the highest probe AUC as the intervention layer for each model.

Table~\ref{tab:layer_selection} summarizes the selected intervention layer and corresponding PCA configuration for each model.
All PCA representations retain more than $90\%$ of the original variance.

\begin{table}[!t]
    \centering
    \setlength{\tabcolsep}{3pt}
    \renewcommand{\arraystretch}{1.1}
    \fontsize{9}{10}\selectfont

    \begin{tabular}{lcccc}
        \toprule[1.5pt]
        \textbf{Model}
        & \textbf{Layers}
        & \textbf{Sel. Layer}
        & \textbf{PCA Dim.}
        & \textbf{Ret. Var.} \\
        \midrule

        Qwen3-1.7B
        & 28
        & 20
        & 512
        & 94.61\% \\

        Qwen3-4B
        & 36
        & 22
        & 512
        & 94.15\% \\

        Qwen3-8B
        & 36
        & 25
        & 512
        & 92.53\% \\

        Qwen3-14B
        & 40
        & 32
        & 512
        & 91.47\% \\

        \bottomrule[1.5pt]
    \end{tabular}

    \vspace{-0.15cm}
    \caption{
Selected intervention layers and PCA configurations.
``Sel. Layer'' is the layer selected by the highest probe AUC, while ``PCA Dim.'' and ``Ret. Var.'' denote the PCA dimensionality and retained variance at that layer, respectively.
}
    \label{tab:layer_selection}
    \vspace{-0.25cm}
\end{table}

\subsection{Prefix State and Latent Direction Clustering}
\label{app:clustering_details}

\begin{figure*}[t]
    \centering
    \includegraphics[width=0.24\textwidth]{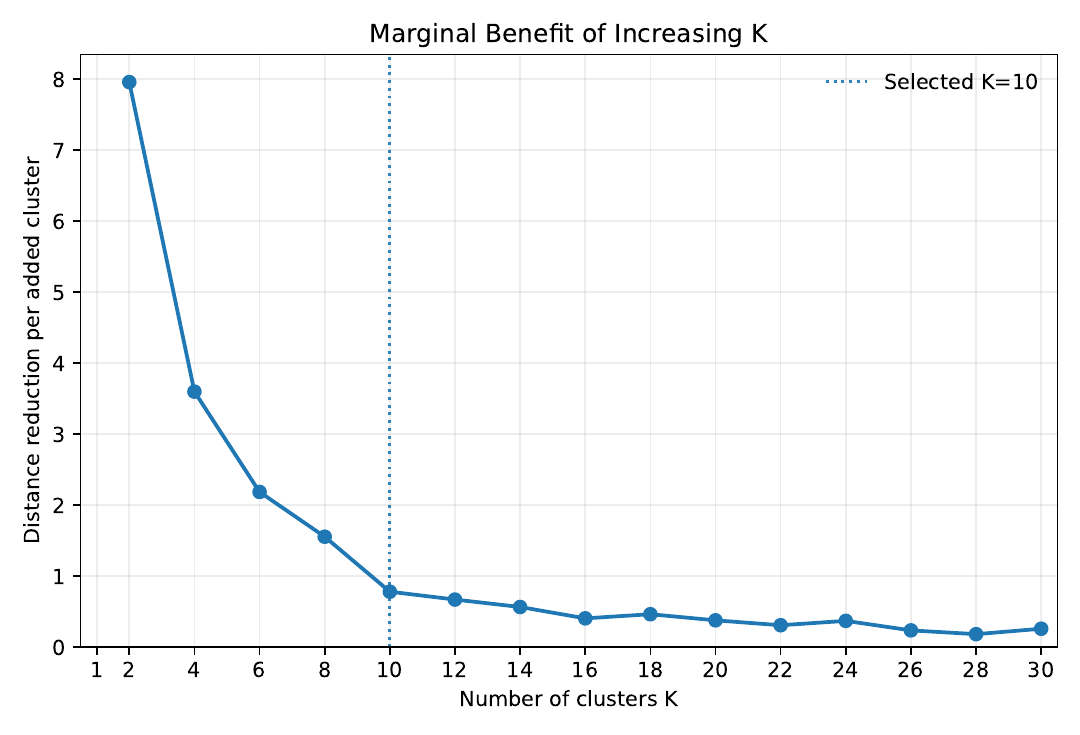}
    \includegraphics[width=0.24\textwidth]{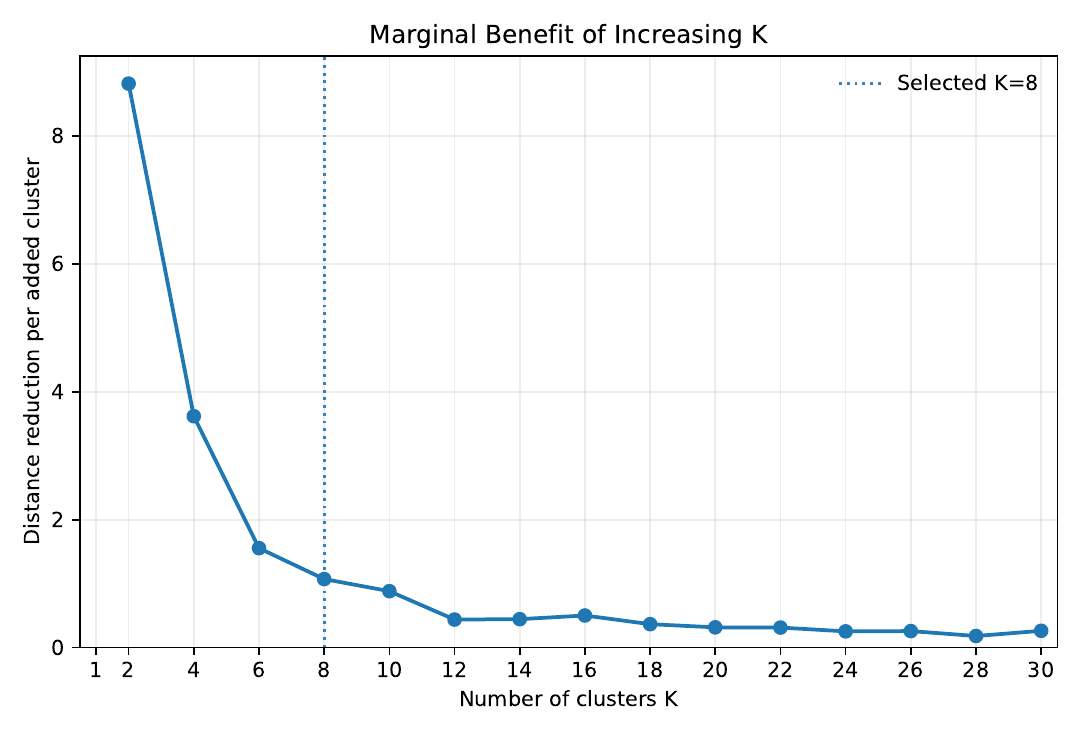}
    \includegraphics[width=0.24\textwidth]{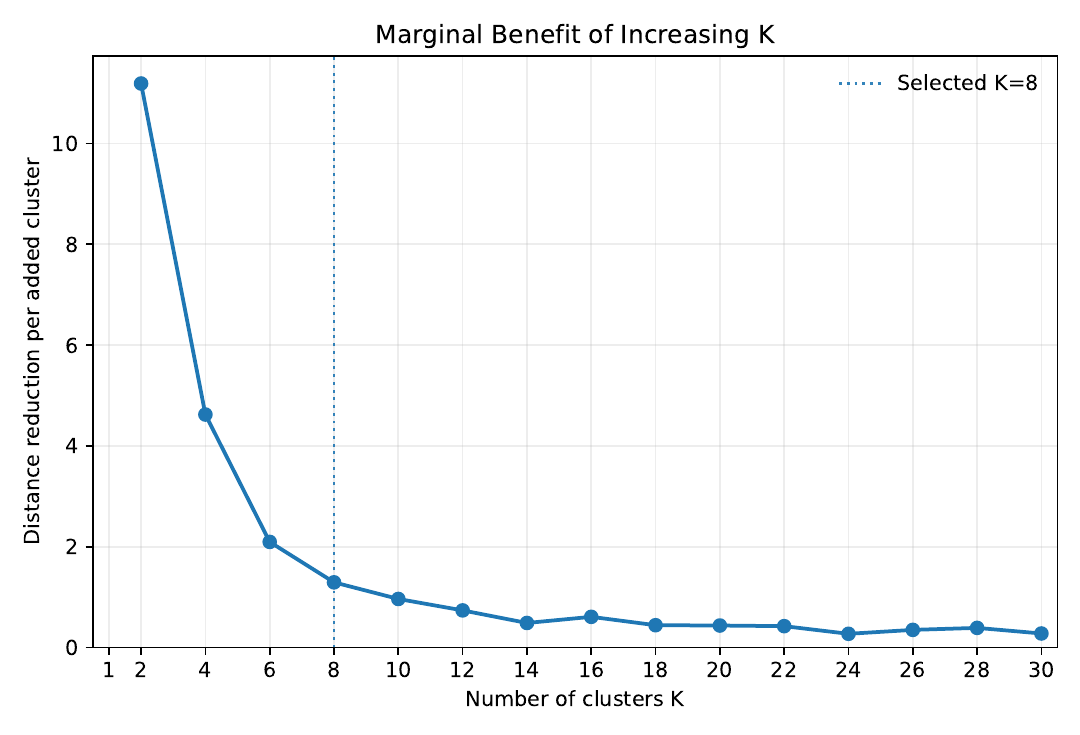}
    \includegraphics[width=0.24\textwidth]{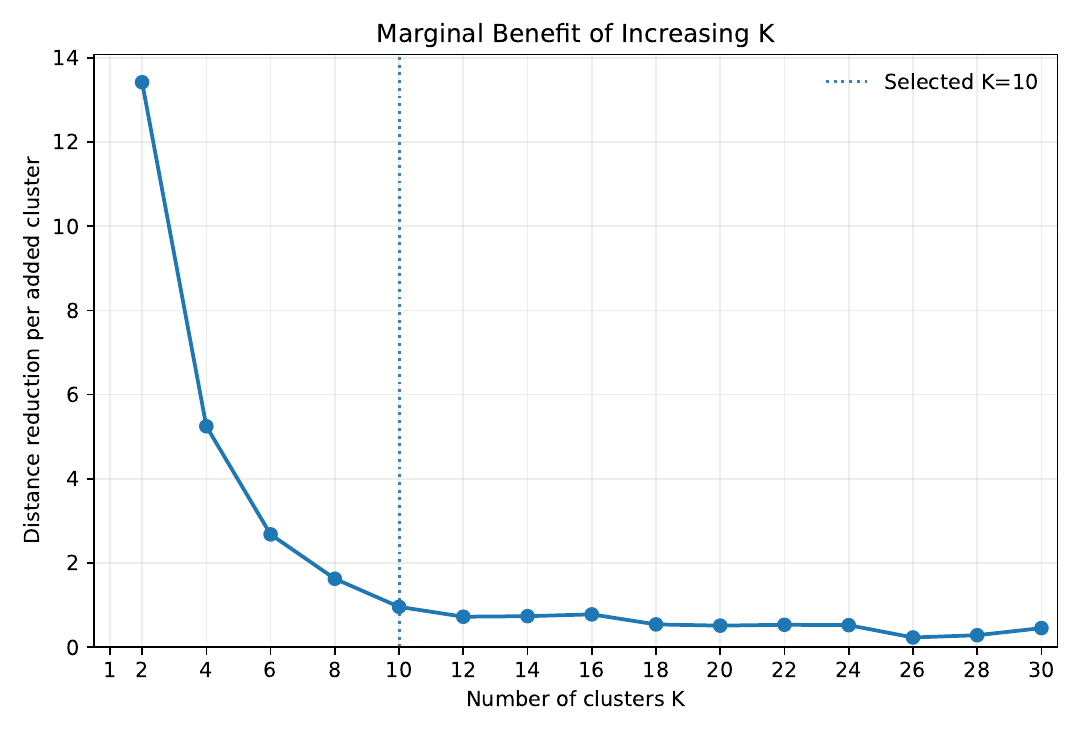}
    \caption{Marginal reduction in mean within-cluster distance per added cluster as the number of prefix-state clusters increases. The vertical dotted line indicates the selected number of regions $C$ for each model. From left to right: Qwen3-1.7B, Qwen3-4B, Qwen3-8B, and Qwen3-14B.}
    \label{fig:prefix_cluster_selection}
\end{figure*}
\nocite{zhang2026beyond,chen2026residual}

\paragraph{Clustering in PCA Space.}
Before clustering, we apply PCA to both prefix states $s_p$ and prefix-specific latent directions $\delta_{p,a}$, and perform clustering on their PCA representations.
For each model, the PCA dimensionality is set to the same value used in the linear probing analysis for automatic intervention layer selection, as reported in Table~\ref{tab:layer_selection}.

\paragraph{Cluster Number Determination.}
We determine the number of clusters based on the marginal reduction in within-cluster distance as $K$ increases. Let $D(K)$ denote the mean Euclidean distance between each PCA representation and its assigned cluster center under $K$ clusters. 
For consecutive candidate cluster numbers $K_{j-1}$ and $K_j$, we compute
\begin{equation}
    R(K_j)
    =
    \frac{
        D(K_{j-1})-D(K_j)
    }{
        K_j-K_{j-1}
    },
    \label{eq:marginal_cluster_reduction}
\end{equation}
where $R(K_j)$ measures the average reduction in within-cluster distance per added cluster when increasing $K$ from $K_{j-1}$ to $K_j$. For example, when the candidate values include $K=2$ and $K=4$, the value at $K=4$ is
$\left(D(2)-D(4)\right)/2$.
\nocite{zhang2025find,su2026massive}

We select $K$ when the marginal reduction enters a stable low-gain region, where further increasing the number of clusters provides limited additional reduction. For prefix-state clustering, the selected $K$ determines the number of prefix regions $C$. We show the corresponding marginal-reduction curves for different models in Fig.~\ref{fig:prefix_cluster_selection}. The same procedure is applied to the latent directions within each region $\mathcal{S}_c$ to determine $M_c$. Since this produces many region-specific curves, we report the resulting $M_c$ values directly in Table~\ref{tab:direction_cluster_numbers}.

Note that PCA is used only to determine the cluster assignments. After clustering, we compute the centroids from the original representations rather than their PCA projections. Specifically, the region centroid $\bar{s}_c$ is computed from the original prefix states $s_p$ assigned to $\mathcal{S}_c$, while each direction centroid $\bar{d}_{c,m}$ is computed from the original latent directions $\delta_{p,a}$ assigned to $\mathcal{D}_{c,m}$, following Eqs.~\ref{eq:state_centroid} and~\ref{eq:steering_vector}. The resulting region centroids and steering vectors therefore remain in the original hidden-state space.

\begin{table}[!h]
    \centering
    \setlength\tabcolsep{2pt}
    \fontsize{10}{13}\selectfont
    \begin{tabular}{lc}
        \toprule
        \textbf{Model} & \textbf{Region-wise Direction Clusters $M_c$} \\
        \midrule
        % Qwen3-0.6B  & $[\cdots]$ \\
        Qwen3-1.7B  & $[6, 6, 4, 6, 4, 8, 6, 8, 4, 6]$ \\
        Qwen3-4B    & $[6, 6, 6, 4, 6, 6, 8, 6]$ \\
        Qwen3-8B    & $[6, 4, 6, 6, 4, 8, 6, 6]$ \\
        Qwen3-14B   & $[6, 6, 4, 6, 8, 4, 6, 6, 4, 8]$ \\
        \bottomrule
    \end{tabular}
    \caption{Latent-direction cluster counts ($M_c$) across prefix regions for each model.}
    \label{tab:direction_cluster_numbers}
\end{table}

\subsection{Entropy-Adaptive Steering Strength Adjustment}
\label{app:entropy_adaptive_strength}
In the main text, we describe Entropy-Adaptive Strength using the entropy percentile $F_H(H_t)$ for clarity. In practice, however, SPS only applies steering when the current transition passes the Uncertainty Gate, i.e., when $H_t>\tau_H$. Therefore, directly using $F_H(H_t)$ as the steering strength would restrict the effective strength range to the upper tail of the entropy distribution. For example, if $\tau_H$ is set to the 80th percentile, all triggered transitions would have $F_H(H_t)>0.8$, resulting in steering strengths concentrated within a narrow range.
\nocite{xiong2026mmformalizer,chang2025treereview,chen-etal-2025-guilomo}

To provide a wider range of steering strengths among the triggered transitions, we rescale the entropy percentile above the uncertainty threshold and map it to $[\alpha_{\min},1]$:
\begin{equation}
    \alpha_t
    =
    \alpha_{\min}
    +
    (1-\alpha_{\min})
    \frac{
        F_H(H_t)-F_H(\tau_H)
    }{
        1-F_H(\tau_H)
    }.
    \label{eq:entropy_adaptive_strength_impl}
\end{equation}
The normalized term
\begin{equation}
    \frac{
        F_H(H_t)-F_H(\tau_H)
    }{
        1-F_H(\tau_H)
    }
    \in [0,1]
\end{equation}
measures the relative uncertainty of the current transition within the portion of the entropy distribution that passes the Uncertainty Gate. Consequently, $\alpha_t\in[\alpha_{\min},1]$, where $\alpha_{\min}$ determines the minimum steering strength applied to a triggered transition. We set $\alpha_{\min}=0.5$ in our experiments.

As a result, a transition whose entropy is close to the uncertainty threshold receives a steering strength close to $\alpha_{\min}$, while increasingly uncertain transitions receive progressively stronger steering, up to $1.0$. This design preserves the monotonic relationship between transition uncertainty and steering strength while providing sufficient variation among the transitions at which SPS intervenes.

\section{Baseline Details}
\label{app:baseline_details}

\paragraph{Chain-of-Thought (CoT).}
CoT~\citep{cot} serves as the standard inference baseline, where the original reasoning model generates a complete reasoning trajectory without additional exploration mechanisms. We use the standard sampling settings described in \S\ref{para:implementation_details}. Unless otherwise specified, all methods in our experiments use the following prompt format for generation.

\begin{tcolorbox}[
    enhanced,
    breakable,
    colback=gray!5,
    colframe=black,
    colbacktitle=black,
    coltitle=white,
    title=\textbf{Default Generation Prompt},
    fonttitle=\bfseries,
    boxrule=0.8pt,
    arc=2pt,
    left=6pt,
    right=6pt,
    top=6pt,
    bottom=6pt
]
\small
\ttfamily
\textless|im\_start|\textgreater user\\
\{input\}\\
Please reason step by step, and put your final answer within \textbackslash boxed\{\{\}\}.\textless|im\_end|\textgreater\\
\textless|im\_start|\textgreater assistant
\end{tcolorbox}

\paragraph{DIPPER.}

DIPPER~\citep{dipper} promotes reasoning diversity by conditioning different inference rollouts on different reasoning instructions. We use the seven fixed reasoning prompts provided in the original work. To estimate the effectiveness of each prompt, we construct a held-out calibration set by randomly sampling 200 Level-8/9 problems from DeepMath-103K~\citep{deepmath}, after filtering out problems with yes/no answers. For each evaluated model, we independently evaluate the seven prompts on this fixed calibration set and use their corresponding accuracies as sampling weights. Specifically, for the $i$-th prompt $w_i$, its sampling probability is defined as
\[
p(w_i)
=
\frac{u(w_i)}
{\sum_{j=1}^{7} u(w_j)},
\]
where $u(w_i)$ denotes the accuracy obtained using prompt $w_i$ on the calibration set. During evaluation, each rollout independently samples a reasoning prompt with replacement according to this distribution. 
The seven reasoning prompts are:
\begin{tcolorbox}[
    enhanced,
    breakable,
    colback=gray!5,
    colframe=black,
    boxrule=0.8pt,
    arc=2pt,
    left=6pt,
    right=6pt,
    top=6pt,
    bottom=6pt
]
\small

\textbf{1.} \textit{Let's think step-by-step to find the answer}

\vspace{0.4em}
\textbf{2.} \textit{Reflect on the question carefully before answering}

\vspace{0.4em}
\textbf{3.} \textit{Rephrase the question in your own words before responding}

\vspace{0.4em}
\textbf{4.} \textit{Actively reason through the question and answer each part systematically}

\vspace{0.4em}
\textbf{5.} \textit{Answer this question as a scientist would}

\vspace{0.4em}
\textbf{6.} \textit{Eliminate the obviously incorrect answers first and then choose the most likely correct answer}

\vspace{0.4em}
\textbf{7.} \textit{Analyze the context of the question and use relevant information to derive the answer}

\end{tcolorbox}

For each rollout, the selected reasoning prompt is inserted into the following template. 

\begin{tcolorbox}[
    enhanced,
    breakable,
    colback=gray!5,
    colframe=black,
    colbacktitle=black,
    coltitle=white,
    title=\textbf{DIPPER Generation Prompt},
    fonttitle=\bfseries,
    boxrule=0.8pt,
    arc=2pt,
    left=6pt,
    right=6pt,
    top=6pt,
    bottom=6pt
]
\small
\ttfamily
\textless|im\_start|\textgreater user\\
\{input\}\\
\{selected reasoning prompt\}, and put your final answer within \textbackslash boxed\{\{\}\}.\textless|im\_end|\textgreater\\
\textless|im\_start|\textgreater assistant
\end{tcolorbox}

\paragraph{WiSE-FT.}
WiSE-FT~\citep{wiseft} improves reasoning exploration through weight interpolation between a base model and its corresponding trained model. Let $\theta_{\mathrm{base}}$ and $\theta_{\mathrm{trained}}$ denote the parameters of the model before and after post-training, respectively. The interpolated model is constructed as
\begin{equation}
    \theta_{\mathrm{WiSE}}
    =
    (1-\delta)\theta_{\mathrm{base}}
    +
    \delta\theta_{\mathrm{trained}},
\end{equation}
where $\delta$ controls the contribution of the trained model. 
In our experiments, we interpolate each Qwen3 base model with its corresponding post-trained reasoning model. 
We select $\delta$ from $\{0.5,0.7,0.9\}$ using 200 randomly sampled non-Yes/No Level-8/9 problems from DeepMath-103K~\citep{deepmath}, and report the best-performing setting.

\paragraph{Entropy-Triggered Temperature (ETT).}
Inspired by the decoding analysis in~\citet{28rule}, ETT assigns different temperatures according to token entropy:
\begin{equation}
    T_t
    =
    \begin{cases}
        T_{\mathrm{high}}, & H_t > \tau_{\mathrm{ETT}}, \\
        T_{\mathrm{low}},  & \text{otherwise}.
    \end{cases}
\end{equation}
The original work does not assign a specific name to this decoding strategy; we refer to it as \emph{Entropy-Triggered Temperature (ETT)} for convenience. Following the original setting, $\tau_{\mathrm{ETT}}$ is determined by the 80th percentile of the token entropy distribution. We set $T_{\mathrm{low}}=1.0$ and $T_{\mathrm{high}}=2.0$, where $T_{\mathrm{high}}=2.0$ gives the best average AIME performance in the temperature sweep reported by~\citet{28rule}.

\paragraph{Latent Exploration Decoding (LED).}
LED~\citep{led} restores exploration by exploiting higher-entropy intermediate-layer posteriors during decoding. It restricts exploration to final-layer top-$k$ token candidates, cumulatively aggregates their posteriors across multiple intermediate layers, and selects the aggregated posterior with the highest entropy for exploration. Following the official implementation, we set the exploration depth to $d=8$ and temperature to $0.6$. We disable top-$p$ sampling and use top-$k=8$ for LED. We also follow the default implementation by not applying the final LayerNorm to intermediate-layer hidden states.

\section{Evaluation Benchmark Details}
\label{app:benchmark_details}
We evaluate SPS on nine benchmarks spanning mathematics, science, coding, and commonsense reasoning.

\paragraph{MATH-500.}
MATH-500~\citep{math500} is a representative subset of the MATH benchmark containing 500 problems across diverse mathematical subjects, including algebra, geometry, number theory, and counting and probability.

\paragraph{AIME 2024.}
AIME 2024~\citep{aime24} contains 30 problems from the 2024 American Invitational Mathematics Examination (AIME I and AIME II). The problems require advanced competition-level mathematical reasoning and have integer-valued final answers.

\paragraph{AIME 2025.}
AIME 2025~\citep{aime25} contains 30 problems from the 2025 AIME I and AIME II. It follows the same competition format as AIME 2024 and evaluates advanced mathematical problem solving.

\paragraph{Minerva-Math.}
Minerva-Math~\citep{Minerva} contains 272 mathematical problems covering a broad range of quantitative reasoning topics. The benchmark includes free-form mathematical questions that require multi-step reasoning and answer derivation.

\paragraph{OlympiadBench.}
OlympiadBench~\citep{olympiadbench} is a challenging benchmark constructed from mathematics and physics problems collected from international Olympiads and high-level entrance examinations. We use its text-only mathematics subset containing 675 problems.

\paragraph{HMMT 2025.}
HMMT 2025~\citep{hmmt25} contains 30 problems from the February 2025 Harvard--MIT Mathematics Tournament. The benchmark consists of competition-level problems covering areas such as algebra, geometry, combinatorics, and number theory.

\paragraph{GPQA-Diamond.}

GPQA-Diamond~\citep{gpqa} comprises 198 graduate-level multiple-choice questions in biology, physics, and chemistry. It is the highest-quality subset of GPQA and is designed to require expert-level scientific reasoning. We use the following prompt for evaluation.

\begin{tcolorbox}[
    enhanced,
    breakable,
    colback=gray!5,
    colframe=black,
    colbacktitle=black,
    coltitle=white,
    title=\textbf{GPQA-Diamond Evaluation Prompt},
    fonttitle=\bfseries,
    boxrule=0.8pt,
    arc=2pt,
    left=6pt,
    right=6pt,
    top=6pt,
    bottom=6pt
]
\small
\ttfamily

Please solve the following multiple-choice question. Think step by step, then end your response with a JSON object whose answer field contains only the choice letter, for example \{"answer": "C"\}.\\
\\
\{question\}

\end{tcolorbox}

\paragraph{LiveCodeBench.}

LiveCodeBench~\citep{livecodebench} is a continuously updated code-generation benchmark designed to reduce data contamination. We use the v5 subset containing 279 problems published between August 2024 and January 2025. Depending on whether starter code is provided, we use the following prompt.

\begin{tcolorbox}[
    enhanced,
    breakable,
    colback=gray!5,
    colframe=black,
    colbacktitle=black,
    coltitle=white,
    title=\textbf{LiveCodeBench Evaluation Prompt (without Starter Code)},
    fonttitle=\bfseries,
    boxrule=0.8pt,
    arc=2pt,
    left=6pt,
    right=6pt,
    top=6pt,
    bottom=6pt
]
\small
\ttfamily

You will be given a question (problem specification) and will generate a correct Python program that matches the specification and passes all tests.\\
\\
Question: \{question\}\\
\\
Read the inputs from stdin, solve the problem, and write the answer to stdout (do not directly test on the sample inputs). Enclose the complete program within delimiters as follows.\\
`python\\
\# YOUR CODE HERE\\
`

\end{tcolorbox}

\paragraph{StrategyQA.}

StrategyQA~\citep{strategyqa} contains 2,290 yes/no questions that require implicit multi-step reasoning. The required reasoning strategy is not explicitly stated in the question, making the benchmark suitable for evaluating commonsense reasoning across diverse topics. We use the following prompt for evaluation.

\begin{tcolorbox}[
    enhanced,
    breakable,
    colback=gray!5,
    colframe=black,
    colbacktitle=black,
    coltitle=white,
    title=\textbf{StrategyQA Evaluation Prompt},
    fonttitle=\bfseries,
    boxrule=0.8pt,
    arc=2pt,
    left=6pt,
    right=6pt,
    top=6pt,
    bottom=6pt
]
\small
\ttfamily

Answer the following StrategyQA question using the supplied facts and your reasoning. Think step by step, then end your response with exactly one JSON object: \{"answer": "Yes"\} or \{"answer": "No"\}.\\
\\
\{question\}

\end{tcolorbox}

\section{Additional Experimental Results}
\label{app:additional_results}
\subsection{Additional Results on Larger Model Scale}
\begin{table*}[!t]
    \centering
    \renewcommand{\arraystretch}{1.15}
    \fontsize{8}{9}\selectfont
    \setlength{\tabcolsep}{3pt}

    \resizebox{\textwidth}{!}{
    \begin{tabular}{l cc cc cc cc cc cc}
        \toprule[1.5pt]
        & \multicolumn{2}{c}{\textbf{MATH}}
        & \multicolumn{2}{c}{\textbf{Minerva}}
        & \multicolumn{2}{c}{\textbf{Olympiad}}
        & \multicolumn{2}{c}{\textbf{AIME24}}
        & \multicolumn{2}{c}{\textbf{AIME25}}
        & \multicolumn{2}{c}{\textbf{HMMT25}} \\
        \cmidrule(lr){2-3}
        \cmidrule(lr){4-5}
        \cmidrule(lr){6-7}
        \cmidrule(lr){8-9}
        \cmidrule(lr){10-11}
        \cmidrule(lr){12-13}

        \textbf{Method}
        & Pass@1 & Pass@4
        & Pass@1 & Pass@4
        & Pass@1 & Pass@4
        & Pass@1 & Pass@4
        & Pass@1 & Pass@4
        & Pass@1 & Pass@4 \\
        \midrule

        \rowcolor{gray!20}
        \multicolumn{13}{c}{\textbf{Qwen3-8B}} \\
        \addlinespace[1pt]

        CoT
        & 95.50 & 97.40
        & 50.92 & 55.88
        & 77.44 & 84.30
        & 71.67 & 80.00
        & 66.67 & 76.67
        & 43.33 & 56.67 \\

        DIPPER
        & 95.85 & 98.00
        & 51.10 & 56.25
        & 77.41 & 84.44
        & 70.83 & 80.00
        & 67.50 & 76.67
        & 41.67 & 56.67 \\

        WiSE-FT
        & 95.30 & 97.20
        & 50.55 & 55.51
        & 78.04 & 85.33
        & 70.00 & 83.33
        & 65.00 & 73.33
        & 40.83 & 60.00 \\

        ETT
        & 95.75 & 97.80
        & 51.65 & 56.99
        & 77.81 & 85.04
        & 74.17 & 83.33
        & 69.17 & 80.00
        & 45.83 & 63.33 \\

        LED
        & 95.65 & 97.60
        & 52.02 & 57.72
        & 77.93 & 85.19
        & 75.00 & 86.67
        & 70.83 & 83.33
        & 46.67 & 63.33 \\

        \textbf{SPS (Ours)}
        & \textbf{96.10} & \textbf{98.40}
        & \textbf{52.67} & \textbf{59.19}
        & \textbf{79.04} & \textbf{87.11}
        & \textbf{77.50} & \textbf{90.00}
        & \textbf{71.67} & \textbf{86.67}
        & \textbf{49.17} & \textbf{70.00} \\

        \midrule

        \rowcolor{gray!20}
        \multicolumn{13}{c}{\textbf{Qwen3-14B}} \\
        \addlinespace[1pt]

        CoT
        & 96.40 & 97.60
        & 52.39 & 57.72
        & 79.67 & 86.07
        & 79.17 & 86.67
        & 68.33 & 83.33
        & 47.50 & 60.00 \\

        DIPPER
        & 96.50 & 97.80
        & 52.76 & 58.82
        & 79.85 & 86.37
        & 80.00 & 86.67
        & 69.17 & 83.33
        & 46.67 & 56.67 \\

        WiSE-FT
        & 96.25 & 97.80
        & 51.93 & 56.99
        & 79.37 & 86.67
        & 77.50 & 83.33
        & 65.83 & 80.00
        & 45.83 & 63.33 \\

        ETT
        & 96.80 & 98.20
        & 53.22 & 59.19
        & 80.30 & 87.11
        & 81.67 & 90.00
        & 70.83 & 86.67
        & 50.00 & 63.33 \\

        LED
        & 96.65 & 98.00
        & 53.40 & 59.56
        & 80.15 & 86.96
        & 82.50 & 90.00
        & 71.67 & 86.67
        & 50.83 & 66.67 \\

        \textbf{SPS (Ours)}
        & \textbf{96.95} & \textbf{98.80}
        & \textbf{54.32} & \textbf{61.76}
        & \textbf{81.11} & \textbf{89.04}
        & \textbf{85.83} & \textbf{93.33}
        & \textbf{75.83} & \textbf{90.00}
        & \textbf{53.33} & \textbf{73.33} \\

        \bottomrule[1.5pt]
    \end{tabular}
    }

    \vspace{-0.15cm}
    \caption{
    Results of SPS and baseline methods on Qwen3-8B and Qwen3-14B across six mathematical reasoning benchmarks.
    See Appendix~\ref{app:baseline_details} for the details of baseline methods.
    }
    \label{tab:main_results_large}
    \vspace{-0.2cm}
\end{table*}

We report the detailed results of SPS and all baseline methods on Qwen3-8B and Qwen3-14B across six mathematical reasoning benchmarks in Table~\ref{tab:main_results_large}.
SPS consistently achieves the best performance across both model scales.
For example, on Qwen3-8B, SPS improves Pass@1 on HMMT25 by 5.36\% over LED.
On Qwen3-14B, SPS improves Pass@4 on OlympiadBench by 2.22\% over ETT.
These results further confirm that the effectiveness of SPS is maintained as model scale increases.

\subsection{Additional Results on Cross-Domain Benchmarks}

Table~\ref{tab:generalization_detailed} reports the detailed Pass@1 and Pass@4 results of SPS and baseline methods on GPQA-Diamond, StrategyQA, and LiveCodeBench across four model scales.
For example, on GPQA-Diamond with Qwen3-1.7B, SPS achieves relative improvements of 6.19\% in Pass@1 and 5.64\% in Pass@4 over ETT.
On LiveCodeBench with Qwen3-4B, SPS improves over ETT by 2.46\% and 2.67\% in Pass@1 and Pass@4, respectively.
These results further show that SPS generalizes effectively beyond the mathematical domain.

\subsection{Additional Ablation Results}
\label{app:ablation_details}

We provide additional details of the four ablation variants used in our experiments.

\paragraph{SPS w/o Direction Bank (DB).}
For each state region, we directly sum all prefix-specific progress-guided directions into a single state-specific steering vector, instead of clustering them into multiple steering vectors to construct a Direction Bank.

\paragraph{SPS w/o State Matching (SM).}
We shuffle the state-region assignments during online inference, such that the selected Direction Bank is no longer matched to the current prefix state.
A steering vector is then sampled from the assigned Direction Bank as in the full SPS framework.

\begin{table}[!t]
    \centering
    \renewcommand{\arraystretch}{1.15}
    \fontsize{8}{9}\selectfont
    \setlength{\tabcolsep}{4pt}

    \resizebox{\columnwidth}{!}{
    \begin{tabular}{l cc cc cc}
        \toprule[1.5pt]
        & \multicolumn{2}{c}{\textbf{GPQA-Diamond}}
        & \multicolumn{2}{c}{\textbf{StrategyQA}}
        & \multicolumn{2}{c}{\textbf{LiveCodeBench}} \\
        \cmidrule(lr){2-3}
        \cmidrule(lr){4-5}
        \cmidrule(lr){6-7}

        \textbf{Method}
        & Pass@1 & Pass@4
        & Pass@1 & Pass@4
        & Pass@1 & Pass@4 \\
        \midrule

        \rowcolor{gray!20}
        \multicolumn{7}{c}{\textbf{Qwen3-1.7B}} \\
        \addlinespace[1pt]

        CoT
        & 37.12 & 59.60
        & 88.59 & 94.24
        & 35.39 & 47.31 \\

        DIPPER
        & 37.63 & 60.61
        & 88.36 & 94.41
        & 36.02 & 48.03 \\

        WiSE-FT
        & 36.87 & 61.62
        & 88.30 & 95.11
        & 35.13 & 47.67 \\

        ETT
        & 38.76 & 62.63
        & 88.77 & 95.20
        & 36.47 & 49.10 \\

        LED
        & 39.27 & 63.64
        & 88.93 & 95.46
        & 36.92 & 49.82 \\

        \textbf{SPS (Ours)}
        & \textbf{41.16} & \textbf{66.16}
        & \textbf{89.36} & \textbf{95.50}
        & \textbf{37.90} & \textbf{50.90} \\

        \midrule

        \rowcolor{gray!20}
        \multicolumn{7}{c}{\textbf{Qwen3-4B}} \\
        \addlinespace[1pt]

        CoT
        & 54.04 & 70.71
        & 93.79 & 96.46
        & 53.94 & 65.23 \\

        DIPPER
        & 53.79 & 70.71
        & 93.28 & 95.59
        & 53.67 & 64.87 \\

        WiSE-FT
        & 53.41 & 72.22
        & 93.36 & 96.24
        & 53.23 & 65.95 \\

        ETT
        & 55.18 & 73.23
        & 93.77 & 96.29
        & 54.48 & 67.03 \\

        LED
        & 55.81 & 74.24
        & 93.90 & 96.51
        & 54.84 & 67.38 \\

        \textbf{SPS (Ours)}
        & \textbf{57.58} & \textbf{76.26}
        & \textbf{94.13} & \textbf{96.55}
        & \textbf{55.82} & \textbf{68.82} \\

        \midrule

        \rowcolor{gray!20}
        \multicolumn{7}{c}{\textbf{Qwen3-8B}} \\
        \addlinespace[1pt]

        CoT
        & 57.95 & 72.22
        & 94.20 & 96.90
        & 58.33 & 65.59 \\

        DIPPER
        & 58.46 & 72.73
        & 94.16 & 96.94
        & 58.87 & 66.31 \\

        WiSE-FT
        & 57.58 & 75.25
        & 94.30 & 97.42
        & 57.89 & 67.38 \\

        ETT
        & 59.47 & 74.75
        & 94.43 & 97.21
        & 59.32 & 67.38 \\

        LED
        & 60.10 & 75.76
        & 94.55 & 97.07
        & 59.86 & 68.82 \\

        \textbf{SPS (Ours)}
        & \textbf{61.36} & \textbf{77.27}
        & \textbf{94.76} & \textbf{97.64}
        & \textbf{60.66} & \textbf{69.53} \\

        \midrule

        \rowcolor{gray!20}
        \multicolumn{7}{c}{\textbf{Qwen3-14B}} \\
        \addlinespace[1pt]

        CoT
        & 61.24 & 74.24
        & 94.69 & 96.94
        & 62.28 & 70.97 \\

        DIPPER
        & 61.74 & 74.24
        & 94.75 & 97.21
        & 62.81 & 72.04 \\

        WiSE-FT
        & 60.86 & 75.76
        & 94.63 & 97.69
        & 61.83 & 71.68 \\

        ETT
        & 63.13 & 77.27
        & 95.15 & 97.90
        & 63.80 & 73.48 \\

        LED
        & 63.01 & 76.26
        & 94.72 & 97.42
        & 63.53 & 72.76 \\

        \textbf{SPS (Ours)}
        & \textbf{64.90} & \textbf{79.29}
        & \textbf{95.20} & \textbf{97.95}
        & \textbf{64.87} & \textbf{74.19} \\

        \bottomrule[1.5pt]
    \end{tabular}
    }

    \vspace{-0.2cm}
    \caption{
    Detailed results of SPS and baselines across GPQA-Diamond, StrategyQA, and LiveCodeBench.
    }
    \label{tab:generalization_detailed}
    \vspace{-0.6cm}
\end{table}

\paragraph{SPS w/o Uncertainty Gate (UG).}
We apply intervention at every reasoning-step boundary rather than only at high-uncertainty transitions.
Entropy-adaptive steering strength is retained, with the strength determined by the entropy at each intervention position.

\begin{figure}[!t]
    \centering
    \centerline{
        \includegraphics[width=\columnwidth]{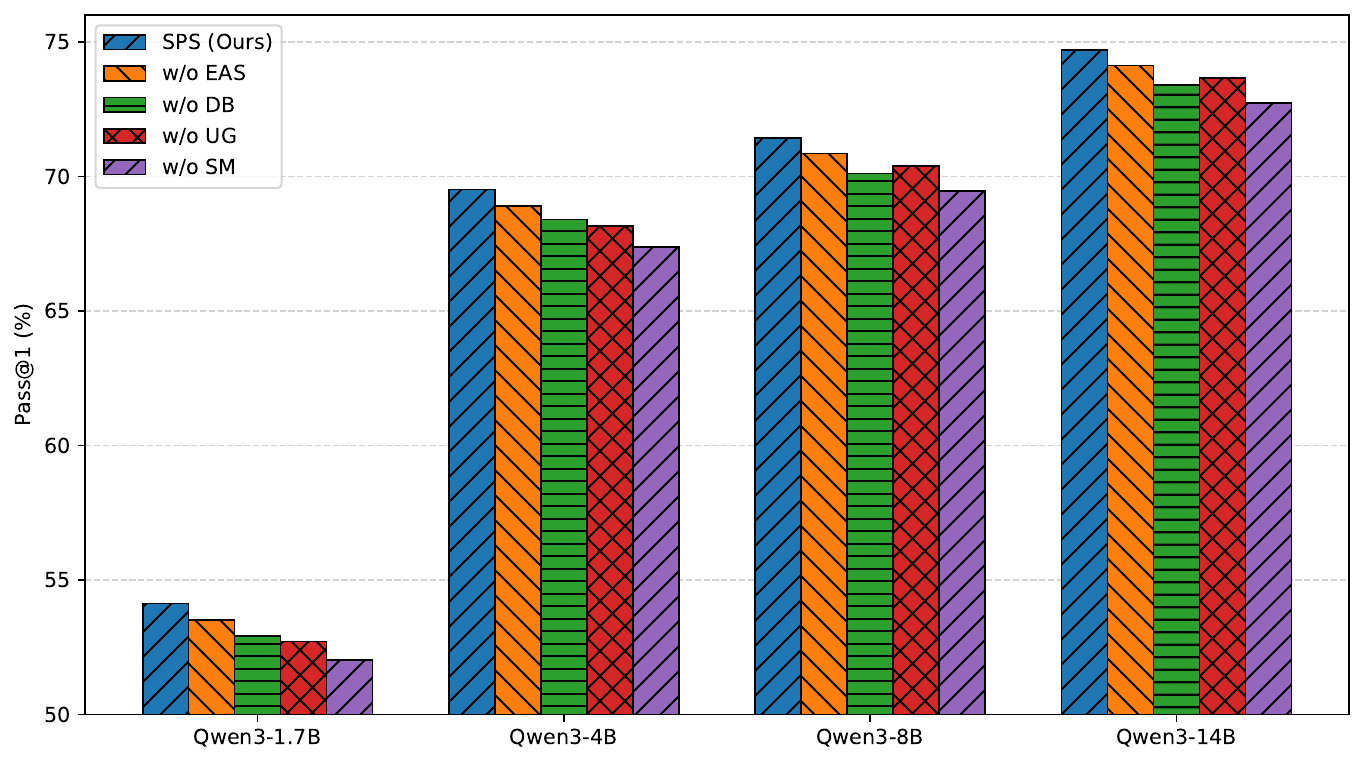}
    }
    \caption{
    Average Pass@1 of SPS and its ablation variants across nine benchmarks.
    }
    \label{fig:ablation_pass1}
\end{figure}

\paragraph{SPS w/o Entropy-Adaptive Strength (EAS).}
We replace the entropy-adaptive strength with a fixed steering strength.
We select the fixed strength from $\{0.5,0.7,0.9\}$ based on its performance on 200 randomly sampled non-Yes/No Level-8/9 problems from DeepMath-103K~\citep{deepmath}, and report the best-performing setting.
Fig.~\ref{fig:ablation_pass1} further reports the average Pass@1 across the nine benchmarks.
Consistent with the Pass@4 results, all ablation variants underperform the complete SPS framework, with removing State Matching causing the most pronounced overall degradation.

\section{Additional Analyses}
\label{app:additional_analysis}

\subsection{Efficiency Analysis}
\label{app:efficiency}
We evaluate the inference efficiency of SPS on AIME24 using Qwen3-1.7B and Qwen3-4B on a single NVIDIA H20 GPU.
We report tokens per second (TPS), time per request (TPR), and the average number of generated tokens (\#Tokens), where TPS measures output throughput and TPR denotes the total wall-clock inference time per request.

Table~\ref{tab:efficiency_analysis} shows that SPS introduces only minor token-generation overhead.
Compared with standard CoT, TPS decreases by only 1.90\% and 1.79\% on Qwen3-1.7B and Qwen3-4B, respectively, due to the additional entropy-based gating, state matching, and lightweight vector addition.
Meanwhile, SPS reduces the average number of generated tokens by 3.60\% and 3.10\%, respectively.
Since steering is applied only at high-uncertainty transitions, SPS preserves the overall generation process while reducing unnecessary reasoning.
The shorter sequences therefore offset the steering overhead, reducing TPR from 12.58\,s to 12.36\,s on Qwen3-1.7B and from 15.48\,s to 15.27\,s on Qwen3-4B.

The additional memory overhead is also negligible.
For Qwen3-4B, the direction bank contains around 50 BF16 steering vectors across all state regions, requiring less than 300\,KB in total ($\sim$5\,KB per vector).
Overall, SPS improves reasoning exploration with only minor inference overhead.

\begin{table}[!t]
    \centering
    \setlength{\tabcolsep}{6pt}
    \renewcommand{\arraystretch}{1.08}
    \fontsize{9}{10.5}\selectfont

    \begin{tabular}{lccc}
        \toprule[1.5pt]
        \textbf{Method}
        & \textbf{TPS}
        & \textbf{TPR (s)}
        & \textbf{\#Tokens} \\
        \midrule

        \rowcolor{gray!20}
        \multicolumn{4}{c}{\textbf{Qwen3-1.7B}} \\
        \addlinespace[1pt]

        CoT      & 1264.0 & 12.58 & 15898.7 \\
        DIPPER   & 1260.8 & 12.65 & 15954.3 \\
        WiSE-FT  & 1262.7 & 12.52 & 15803.3 \\
        ETT      & 1246.9 & 12.86 & 16033.8 \\
        LED      & 1226.1 & 12.99 & 15930.5 \\
        SPS (Ours) & 1240.0 & 12.36 & 15326.3 \\

        \midrule

        \rowcolor{gray!20}
        \multicolumn{4}{c}{\textbf{Qwen3-4B}} \\
        \addlinespace[1pt]

        CoT      & 926.3  & 15.48 & 14338.7 \\
        DIPPER   & 924.5  & 15.55 & 14374.6 \\
        WiSE-FT  & 925.4  & 15.42 & 14267.0 \\
        ETT      & 914.8  & 15.70 & 14364.5 \\
        LED      & 900.4  & 15.94 & 14353.1 \\
        SPS (Ours) & 909.7  & 15.27 & 13894.2 \\

        \bottomrule[1.5pt]
    \end{tabular}
    \caption{
    Efficiency analysis on AIME24 using a single NVIDIA H20 GPU.
    TPS, TPR, and \#Tokens denote tokens per second, time per request, and average generated tokens, respectively.
    }
    \vspace{-0.2cm}
    \label{tab:efficiency_analysis}
\end{table}

\subsection{Sensitivity to Calibration Dataset Size for Direction Bank Construction}
\label{app:calibration_dataset_size}
We further study the sensitivity of SPS to the number of calibration prompts used for constructing the Direction Bank.
Specifically, we vary the calibration set size over
$\{50, 100, 150, 200, 250, 300\}$ prompts sampled from DAPO-Math-17K, while keeping all other settings unchanged.
For each model, we report the average Pass@1 and Pass@4 over all nine benchmarks, normalized by the best performance achieved across the evaluated calibration sizes.

\begin{figure}[!h]
    \centering
    \includegraphics[width=\columnwidth]{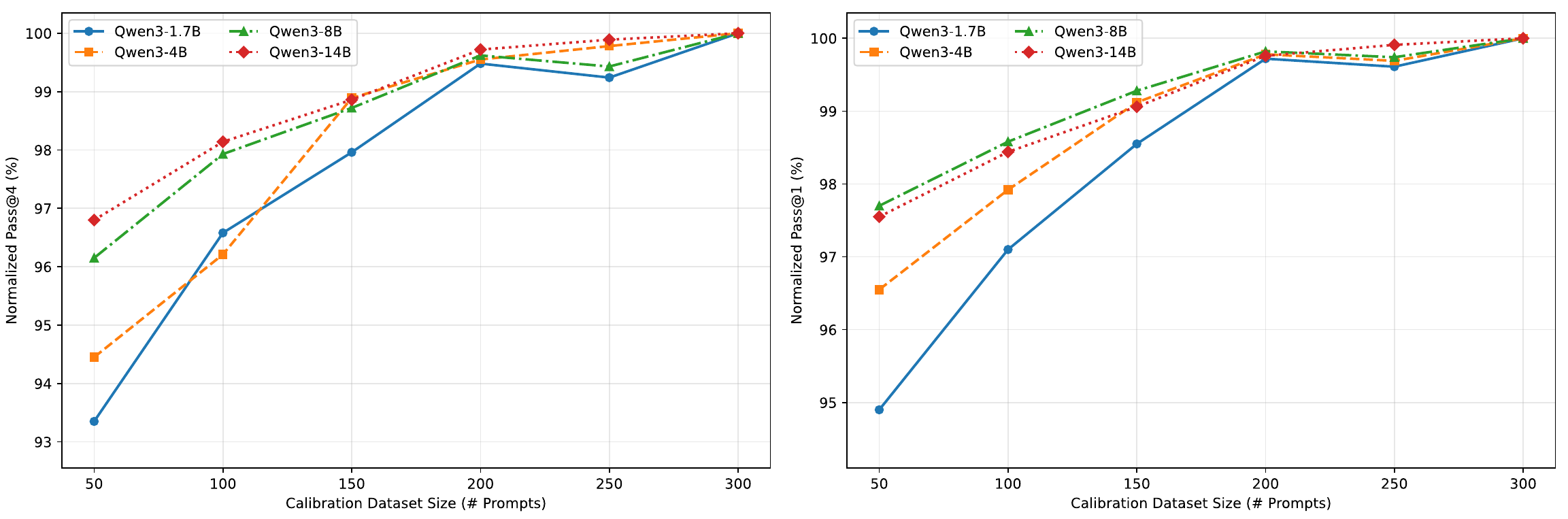}
    \vspace{-0.4cm}
    \caption{
    Sensitivity to the calibration dataset size used for Direction Bank construction.
    Pass@1 and Pass@4 are averaged over nine benchmarks and normalized by each model's best performance across the evaluated calibration sizes.
    }
    \label{fig:calibration_sensitivity}
    \vspace{-0.3cm}
\end{figure}

As shown in Fig.~\ref{fig:calibration_sensitivity}, increasing the calibration set size consistently improves performance in the low-data regime, while the gains gradually diminish as more calibration prompts are added.
The largest improvements generally occur between $50$ and $200$ prompts, after which the curves enter a clear saturation region with only minor fluctuations.
We also observe that Pass@4 is generally more sensitive to the calibration set size than Pass@1, particularly in the low-data regime.
This suggests that broader calibration coverage primarily benefits the diversity and quality of the progress-guided directions available for multi-rollout exploration.
Based on this trade-off between calibration cost and downstream performance, we use $200$ calibration prompts for Direction Bank construction in all experiments.

\newpage
\onecolumn
\subsection{Case Study}
\label{app:case_study}
\begin{figure*}[!h]
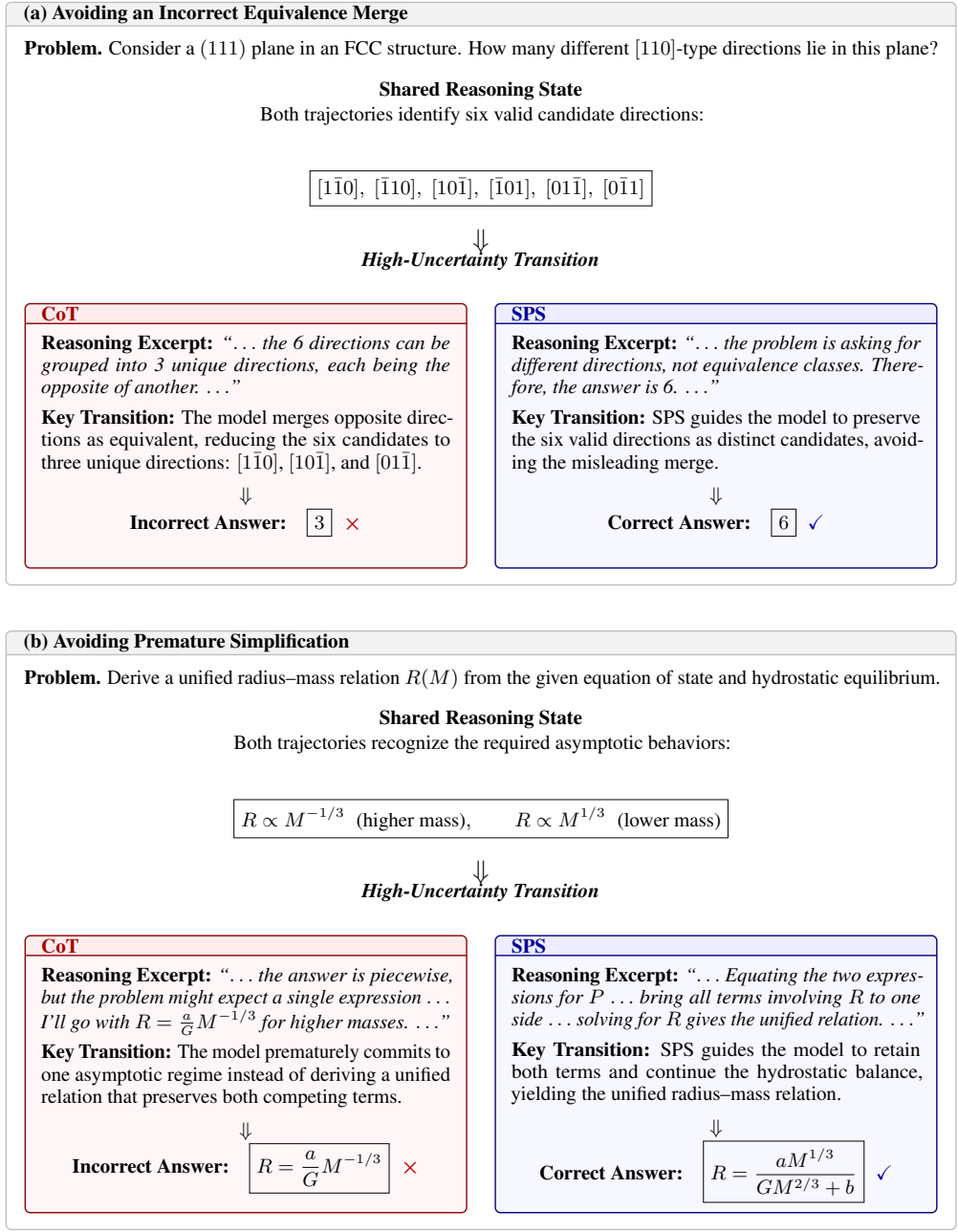

\centering

\resizebox{0.88\textwidth}{!}{%
\begin{minipage}{\textwidth}

% ============================================================
% Case (a)
% ============================================================
\begin{tcolorbox}[
    width=0.94\textwidth,
    colback=white,
    colframe=black!25,
    boxrule=0.6pt,
    arc=2pt,
    left=7pt,
    right=7pt,
    top=5pt,
    bottom=6pt,
    boxsep=1pt,
    title=\textbf{(a) Avoiding an Incorrect Equivalence Merge},
    coltitle=black,
    colbacktitle=black!5,
    fonttitle=\small\bfseries,
    fontupper=\footnotesize
]

\textbf{Problem.}
Consider a $(111)$ plane in an FCC structure.
How many different $[110]$-type directions lie in this plane?

\vspace{0.35em}

\begin{center}
\textbf{Shared Reasoning State}

\vspace{0.15em}

Both trajectories identify six valid candidate directions:

\vspace{0.05em}

\[
\boxed{
[1\bar{1}0],\;
[\bar{1}10],\;
[10\bar{1}],\;
[\bar{1}01],\;
[01\bar{1}],\;
[0\bar{1}1]
}
\]

\vspace{-0.2em}

{\large $\Downarrow$}

\vspace{-0.2em}

\textbf{\textit{High-Uncertainty Transition}}
\end{center}

\vspace{0.2em}

% -------------------- CoT --------------------
\noindent
\begin{minipage}[t]{0.485\linewidth}
\vspace{0pt}

\begin{tcolorbox}[
    height=4.2cm,
    colback=red!3,
    colframe=red!55!black,
    boxrule=0.6pt,
    arc=2pt,
    left=6pt,
    right=6pt,
    top=4pt,
    bottom=4pt,
    boxsep=1pt,
    title=\textbf{CoT},
    coltitle=red!65!black,
    colbacktitle=red!7,
    fonttitle=\footnotesize\bfseries,
    fontupper=\footnotesize,
    valign=top
]

\textbf{Reasoning Excerpt:}
\textit{``$\ldots$ the 6 directions can be grouped into 3 unique directions, each being the opposite of another. $\ldots$''}

\vspace{0.4em}

\textbf{Key Transition:}
The model merges opposite directions as equivalent, reducing the six candidates to three unique directions:
$[1\bar{1}0]$, $[10\bar{1}]$, and $[01\bar{1}]$.

\vfill

\begin{center}
$\Downarrow$

\vspace{0.05em}

\textbf{Incorrect Answer:}\quad
$\boxed{3}\;\;
{\color{red!70!black}\boldsymbol{\times}}$
\end{center}

\end{tcolorbox}
\end{minipage}
\hfill
%
% -------------------- SPS --------------------
\begin{minipage}[t]{0.485\linewidth}
\vspace{0pt}

\begin{tcolorbox}[
    height=4.2cm,
    colback=blue!3,
    colframe=blue!50!black,
    boxrule=0.6pt,
    arc=2pt,
    left=6pt,
    right=6pt,
    top=4pt,
    bottom=4pt,
    boxsep=1pt,
    title=\textbf{SPS},
    coltitle=blue!60!black,
    colbacktitle=blue!7,
    fonttitle=\footnotesize\bfseries,
    fontupper=\footnotesize,
    valign=top
]

\textbf{Reasoning Excerpt:}
\textit{``$\ldots$ the problem is asking for different directions, not equivalence classes. Therefore, the answer is 6. $\ldots$''}

\vspace{0.4em}

\textbf{Key Transition:}
SPS guides the model to preserve the six valid directions as distinct candidates, avoiding the misleading merge.

\vfill

\begin{center}
$\Downarrow$

\vspace{0.05em}

\textbf{Correct Answer:}\quad
$\boxed{6}\;\;
{\color{blue!65!black}\checkmark}$
\end{center}

\end{tcolorbox}
\end{minipage}

\end{tcolorbox}

\vspace{0.6em}

% ============================================================
% Case (b)
% ============================================================
\begin{tcolorbox}[
    width=0.94\textwidth,
    colback=white,
    colframe=black!25,
    boxrule=0.6pt,
    arc=2pt,
    left=7pt,
    right=7pt,
    top=5pt,
    bottom=6pt,
    boxsep=1pt,
    title=\textbf{(b) Avoiding Premature Simplification},
    coltitle=black,
    colbacktitle=black!5,
    fonttitle=\small\bfseries,
    fontupper=\footnotesize
]

\textbf{Problem.}
Derive a unified radius--mass relation $R(M)$ from the given equation of state and hydrostatic equilibrium.

\vspace{0.35em}

\begin{center}
\textbf{Shared Reasoning State}

\vspace{0.15em}

Both trajectories recognize the required asymptotic behaviors:

\vspace{0.05em}

\[
\boxed{
R \propto M^{-1/3}\;\; \text{(higher mass)},
\qquad
R \propto M^{1/3}\;\; \text{(lower mass)}
}
\]

\vspace{-0.2em}

{\large $\Downarrow$}

\vspace{-0.2em}

\textbf{\textit{High-Uncertainty Transition}}
\end{center}

\vspace{0.2em}

% -------------------- CoT --------------------
\noindent
\begin{minipage}[t]{0.485\linewidth}
\vspace{0pt}

\begin{tcolorbox}[
    height=4.4cm,
    colback=red!3,
    colframe=red!55!black,
    boxrule=0.6pt,
    arc=2pt,
    left=6pt,
    right=6pt,
    top=4pt,
    bottom=4pt,
    boxsep=1pt,
    title=\textbf{CoT},
    coltitle=red!65!black,
    colbacktitle=red!7,
    fonttitle=\footnotesize\bfseries,
    fontupper=\footnotesize,
    valign=top
]

\textbf{Reasoning Excerpt:}
\textit{``$\ldots$ the answer is piecewise, but the problem might expect a single expression $\ldots$ I'll go with
$R=\frac{a}{G}M^{-1/3}$ for higher masses. $\ldots$''}

\vspace{0.4em}

\textbf{Key Transition:}
The model prematurely commits to one asymptotic regime instead of deriving a unified relation that preserves both competing terms.

\vfill

\begin{center}
$\Downarrow$

\vspace{0.05em}

\textbf{Incorrect Answer:}\quad
$\boxed{R=\frac{a}{G}M^{-1/3}}\;\;
{\color{red!70!black}\boldsymbol{\times}}$
\end{center}

\end{tcolorbox}
\end{minipage}
\hfill
%
% -------------------- SPS --------------------
\begin{minipage}[t]{0.485\linewidth}
\vspace{0pt}

\begin{tcolorbox}[
    height=4.4cm,
    colback=blue!3,
    colframe=blue!50!black,
    boxrule=0.6pt,
    arc=2pt,
    left=6pt,
    right=6pt,
    top=4pt,
    bottom=4pt,
    boxsep=1pt,
    title=\textbf{SPS},
    coltitle=blue!60!black,
    colbacktitle=blue!7,
    fonttitle=\footnotesize\bfseries,
    fontupper=\footnotesize,
    valign=top
]

\textbf{Reasoning Excerpt:}
\textit{``$\ldots$ Equating the two expressions for $P$ $\ldots$
bring all terms involving $R$ to one side $\ldots$
solving for $R$ gives the unified relation. $\ldots$''}

\vspace{0.4em}

\textbf{Key Transition:}
SPS guides the model to retain both terms and continue the hydrostatic balance, yielding the unified radius--mass relation.

\vfill

\begin{center}
$\Downarrow$

\vspace{0.05em}

\textbf{Correct Answer:}\quad
$\boxed{R=\frac{aM^{1/3}}{GM^{2/3}+b}}\;\;
{\color{blue!65!black}\checkmark}$
\end{center}

\end{tcolorbox}
\end{minipage}

\end{tcolorbox}

\end{minipage}
}

\caption{
Two representative Qwen3-4B cases on Minerva-Math.
At high-uncertainty transitions, standard CoT follows misleading continuations, whereas SPS steers the model toward progress-making reasoning paths that lead to correct solutions.
}
\label{fig:case_studies}

\end{figure*}

We present two representative cases from Qwen3-4B on Minerva-Math in Fig.~\ref{fig:case_studies} to illustrate how SPS affects reasoning at high-uncertainty transitions.
In both cases, standard CoT and SPS reach similar intermediate reasoning states before a critical branching point, but subsequently follow different continuations.
While standard decoding takes a misleading reasoning path, SPS steers the model toward a progress-making continuation that ultimately leads to the correct solution.
These examples provide qualitative evidence that SPS can selectively redirect reasoning at uncertain transitions toward more productive reasoning pathways.

\end{document}